\documentclass[letterpaper]{article}
\usepackage[preprint]{aaai2027}
\usepackage[hyphens]{url}
\usepackage{graphicx}
\usepackage{natbib}
\usepackage{caption}
\usepackage{booktabs}
\usepackage{tabularx}
\usepackage{amsmath}
\usepackage{amssymb}
\usepackage{amsfonts}
\usepackage{algorithm}
\usepackage{algorithmic}
\usepackage{newfloat}
\usepackage{listings}
\usepackage{xcolor}
\usepackage{tikz}
\usetikzlibrary{arrows.meta,backgrounds,fit,positioning,shapes.geometric}
\usepackage{acro}
\usepackage{placeins}

\newcolumntype{Y}{>{\raggedright\arraybackslash}X}
\newcommand{\cpmpy}{CPMpy}
\newcommand{\ortools}{OR-Tools CP-SAT}
\newcommand{\strategy}[1]{\textsc{#1}}
\newtheorem{definition}{Definition}
\newcommand{\dashedmidrule}{%
  \noalign{%
    \vskip1.5pt
    \nointerlineskip
    \hbox to \columnwidth{%
      \leaders\hbox{\rule[0.5ex]{2.2pt}{0.4pt}\hskip2.2pt}\hfill}%
    \nointerlineskip
    \vskip1.5pt}%
}

\definecolor{listingbackground}{gray}{0.97}
\definecolor{listingrule}{gray}{0.76}
\definecolor{listingbreak}{gray}{0.45}
\lstdefinestyle{supplementlisting}{
  basicstyle=\scriptsize\ttfamily,
  backgroundcolor=\color{listingbackground},
  frame=single,
  rulecolor=\color{listingrule},
  framerule=0.35pt,
  framesep=3pt,
  numbers=none,
  showstringspaces=false,
  columns=fullflexible,
  keepspaces=true,
  breaklines=true,
  breakatwhitespace=false,
  postbreak=\mbox{\textcolor{listingbreak}{$\hookrightarrow$}\space},
  tabsize=2,
  captionpos=t,
  abovecaptionskip=0pt,
  belowcaptionskip=0.45em,
  aboveskip=0.5em,
  belowskip=0.6em
}
\DeclareCaptionStyle{ruled}{labelfont=normalfont,labelsep=colon,strut=off}
\floatstyle{ruled}
\newfloat{listing}{tb}{lst}{}
\floatname{listing}{Listing}

\DeclareAcronym{cp}{short=CP,long=Constraint Programming}
\DeclareAcronym{llm}{short=LLM,long=Large Language Model}
\DeclareAcronym{ahd}{short=AHD,long=Automatic Heuristic Design}
\DeclareAcronym{mip}{short=MIP,long=Mixed-Integer Programming}
\DeclareAcronym{csp}{short=CSP,long=Constraint Satisfaction Problem}
\DeclareAcronym{eoh}{short=EoH,long=Evolution of Heuristics}
\DeclareAcronym{pdr}{short=PDR,long=Profile-Diverse Retention}
\DeclareAcronym{mmr}{short=MMR,long=Maximal Marginal Relevance}

\title{LLM-Guided Evolutionary Search for Constraint Model Reformulation to Improve Solver Efficiency}
\author{
Kostis Michailidis\textsuperscript{\rm 1}\thanks{Corresponding author: Kostis Michailidis (\texttt{kostis.michailidis@kuleuven.be}). ORCID iDs: K. Michailidis, 0009-0000-2139-0106; D. Tsouros, 0000-0002-3040-0959; N. Dang, 0000-0002-2693-6953; T. Guns, 0000-0002-2156-2155.},
Dimos Tsouros\textsuperscript{\rm 2},
Nguyen Dang\textsuperscript{\rm 3},
Tias Guns\textsuperscript{\rm 1}
}
\affiliations{
\textsuperscript{\rm 1}DTAI, KU Leuven, Belgium\\
\textsuperscript{\rm 2}University of Western Macedonia, Greece\\
\textsuperscript{\rm 3}University of St Andrews, United Kingdom
}

\begin{document}
\maketitle
\begin{abstract}
Combinatorial problems appear in numerous industrial applications.
A common approach is to formulate these problems as declarative constraint models that can subsequently be compiled to and solved by a range of back-end solvers.
Recent work shows that \acp{llm} can produce \emph{correct} models from natural language, but even a correct model can be expensive to solve because performance remains sensitive to modelling choices.
In this work, we investigate whether \acp{llm} can automate performance-oriented model reformulation.
Inspired by \ac{ahd}, we use an evolutionary framework in which an \ac{llm} proposes candidate reformulations that are verified and benchmarked against the user-defined baseline model.
We compare \ac{ahd}-adapted search strategies that control which prior attempts, instructions, and measured feedback enter each prompt.
Existing retention strategies prioritize recency or performance, but do not explicitly diversify the context.
To address this gap, we introduce \textit{\ac{pdr}}, which applies \ac{mmr} to instance-level runtime vectors to retain behaviourally diverse attempts.
We systematically evaluate the strategies on eight CSPLib problems using validation-based final model selection.
The results show that: (i) iterative reformulation can produce substantial held-out speedups; (ii) strategies that keep the retained context diverse outperform those that retain only recent or the fastest attempts; and (iii) validation-based selection improves the held-out speedup of every strategy.
\end{abstract}

\acresetall

\section{Introduction}

\textit{\Ac{cp}} is a widely used paradigm for solving combinatorial problems,
with applications ranging from scheduling and planning to configuration and verification~\cite{simonis1999building,DBLP:reference/fai/2}.
In the standard \emph{model-and-solve} workflow, a modeller first encodes a problem as a declarative constraint model and then delegates the search for solutions to a state-of-the-art solver.
However, modelling still remains a critical bottleneck, and automating (parts of) it has long been viewed as a "holy grail" for \ac{cp}~\cite{freuder1997holy_grail,freuder2018progress}.

\begin{figure}[t]
\centering
\resizebox{0.95\columnwidth}{!}{\begin{tikzpicture}[
  font=\tiny\sffamily,
  >=Latex,
  compbox/.style={draw, rounded corners=3pt, line width=0.8pt},
  comptitle/.style={font=\tiny\sffamily\bfseries, align=center},
  comptext/.style={font=\tiny\sffamily, align=center},
  flow/.style={->, line width=0.95pt, draw=black!70},
  inner/.style={->, line width=0.75pt, draw=black!75},
  note/.style={font=\tiny\sffamily, text=black!55},
  pics/nn/.style={code={%
    \foreach \ya in {-0.09,0.09} \foreach \yb in {-0.13,0,0.13}
      \draw[orange!70!black, line width=0.4pt] (-0.16,\ya) -- (0,\yb);
    \foreach \yb in {-0.13,0,0.13} \foreach \yc in {-0.09,0.09}
      \draw[orange!70!black, line width=0.4pt] (0,\yb) -- (0.16,\yc);
    \foreach \ya in {-0.09,0.09}
      \fill[orange!25, draw=orange!70!black, line width=0.5pt] (-0.16,\ya) circle (0.034);
    \foreach \yb in {-0.13,0,0.13}
      \fill[orange!25, draw=orange!70!black, line width=0.5pt] (0,\yb) circle (0.034);
    \foreach \yc in {-0.09,0.09}
      \fill[orange!25, draw=orange!70!black, line width=0.5pt] (0.16,\yc) circle (0.034);
  }},
  pics/checkcircle/.style={code={%
    \draw[green!45!black, fill=green!12, line width=0.8pt] (0,0) circle (0.115);
    \draw[green!45!black, line width=1.0pt, line cap=round]
      (-0.05,0.0) -- (-0.01,-0.055) -- (0.075,0.065);
  }},
  pics/cogs/.style={code={%
    \foreach \a in {0,45,...,315}
      \draw[green!45!black, line width=1.3pt, line cap=round]
        (-0.055,0.055) ++(\a:0.06) -- ++(\a:0.062);
    \draw[green!45!black, fill=green!12, line width=0.9pt] (-0.055,0.055) circle (0.085);
    \draw[green!45!black, fill=white, line width=0.7pt] (-0.055,0.055) circle (0.034);
    \foreach \a in {20,80,...,340}
      \draw[green!45!black, line width=1.1pt, line cap=round]
        (0.105,-0.075) ++(\a:0.04) -- ++(\a:0.048);
    \draw[green!45!black, fill=green!12, line width=0.8pt] (0.105,-0.075) circle (0.058);
    \draw[green!45!black, fill=white, line width=0.6pt] (0.105,-0.075) circle (0.023);
  }},
  pics/funnelplain/.style={code={%
    \draw[teal!60!black, fill=teal!20, line width=0.8pt]
      (-0.17,0.13) -- (0.17,0.13) -- (0.04,-0.03) -- (0.04,-0.14)
      -- (-0.04,-0.14) -- (-0.04,-0.03) -- cycle;
  }},
  pics/database/.style={code={%
    \draw[violet!55!black, fill=violet!15, line width=0.8pt]
      (-0.16,0.11) -- (-0.16,-0.12) arc (180:360:0.16 and 0.055) -- (0.16,0.11) -- cycle;
    \draw[violet!55!black, line width=0.6pt] (-0.16,-0.005) arc (180:360:0.16 and 0.055);
    \draw[violet!55!black, fill=violet!30, line width=0.8pt] (0,0.11) ellipse (0.16 and 0.055);
  }},
  pics/funnel/.style={code={%
    \fill[teal!60!black] (-0.10,0.24) circle (0.025)
                         (0.0,0.28)  circle (0.025)
                         (0.10,0.24) circle (0.025);
    \draw[teal!60!black, fill=teal!20, line width=0.8pt]
      (-0.17,0.13) -- (0.17,0.13) -- (0.04,-0.03) -- (0.04,-0.14)
      -- (-0.04,-0.14) -- (-0.04,-0.03) -- cycle;
    \fill[teal!60!black] (0,-0.21) circle (0.025);
  }},
  pics/stopwatch/.style={code={%
    \draw[#1, line width=0.8pt, fill=white] (0,0) circle (0.115);
    \draw[#1, line width=1.1pt, line cap=round] (0,0.115) -- (0,0.175);
    \draw[#1, line width=0.7pt, line cap=round] (0,0.01) -- (0.05,0.06);
  }},
  pics/human/.style={code={%
    \fill[black!40] (0,0.09) circle (0.072);
    \fill[black!40] (-0.115,-0.155)
      .. controls (-0.115,-0.015) and (0.115,-0.015) .. (0.115,-0.155) -- cycle;
  }},
  pics/problemgrid/.style={code={%
    \fill[blue!30] (-0.17,0.055) rectangle (-0.055,0.17);
    \fill[blue!30] (0.055,-0.055) rectangle (0.17,0.055);
    \fill[blue!30] (-0.055,-0.17) rectangle (0.055,-0.055);
    \draw[blue!55!black, line width=0.6pt] (-0.17,-0.17) rectangle (0.17,0.17);
    \draw[blue!55!black, line width=0.5pt]
      (-0.055,-0.17) -- (-0.055,0.17) (0.055,-0.17) -- (0.055,0.17)
      (-0.17,-0.055) -- (0.17,-0.055) (-0.17,0.055) -- (0.17,0.055);
  }}
]

\colorlet{llmcolor}{orange!70!black}
\colorlet{llmfill}{orange!6}
\colorlet{evalcolor}{green!50!black}
\colorlet{evalfill}{green!5}
\colorlet{historycolor}{violet!55!black}
\colorlet{historyfill}{violet!5}
\colorlet{strategycolor}{teal!60!black}
\colorlet{strategyfill}{teal!6}
\colorlet{basecolor}{black!45}
\colorlet{basefill}{black!3}
\colorlet{outcolor}{green!45!black}
\colorlet{outfill}{green!4}

\pic at (0.45,-0.20) {problemgrid};
\pic at (0.85,-0.22) {human};
\node[comptitle] at (0.65,-0.55) {Baseline model $B$};

\node[compbox, draw=basecolor, fill=basefill,
      minimum width=1.0cm, minimum height=2.9cm] (base) at (0.65,-2.225) {};
\pic at (0.65,-1.12) {stopwatch=red!55!black};
\node[note] at (0.65,-1.48) {$\mathcal{D}_{\mathrm{tr}}$};
\fill[red!55!black!55] (0.55,-3.46) rectangle (0.75,-1.72);

\draw[flow, draw=black!55] (1.15,-3.25) -- node[above=0pt, note] {$B$} (1.90,-3.25);

\node[note, anchor=south] at (4.15,-0.19) {reformulation loop (iteration $t$ of $T$)};

\node[compbox, draw=evalcolor, fill=evalfill,
      minimum width=1.90cm, minimum height=1.55cm] (eval) at (2.85,-1.20) {};
\pic at (2.85,-0.725) {cogs};
\node[comptitle, anchor=north] at (2.85,-1.005) {Evaluator};
\node[comptext, anchor=north, text width=1.80cm] at (2.85,-1.30)
  {valid vs.\ $B$?\\ time on $\mathcal{D}_{\mathrm{tr}}$};

\node[compbox, draw=historycolor, fill=historyfill,
      minimum width=1.90cm, minimum height=1.55cm] (history) at (5.45,-1.20) {};
\pic at (5.45,-0.725) {database};
\node[comptitle, anchor=north] at (5.45,-1.005) {History $\mathcal{H}_t$};
\node[comptext, anchor=north] at (5.45,-1.40) {$(\rho_i,m_i,E_i)_{i\le t}$};

\node[compbox, draw=llmcolor, fill=llmfill,
      minimum width=1.90cm, minimum height=1.55cm] (llm) at (2.85,-3.25) {};
\pic at (2.85,-2.775) {nn};
\node[comptitle, anchor=north] at (2.85,-3.055) {LLM};
\node[comptext, anchor=north, text width=1.80cm] at (2.85,-3.35)
  {rationale $\rho_t$,\\ model $m_t$};

\node[compbox, draw=strategycolor, fill=strategyfill, double, double distance=0.8pt,
      minimum width=1.90cm, minimum height=1.55cm] (strategy) at (5.45,-3.25) {};
\pic[scale=0.9] at (5.45,-2.82) {funnel};
\node[comptitle, anchor=north] at (5.45,-3.055) {Strategy $\sigma$};
\node[comptext, anchor=north] at (5.45,-3.45) {$R_{t+1}\subseteq\mathcal{H}_t$};

\draw[inner, draw=llmcolor] (llm.north) --
  node[left=1pt, note, text=llmcolor] {$\rho_t,m_t$} (eval.south);
\draw[inner, draw=evalcolor] (eval.east) --
  node[above=0pt, note, text=green!45!black] {$E_t$} (history.west);
\pic[scale=0.70] at (3.95,-1.45) {checkcircle};
\node[note, text=green!45!black, inner sep=0pt] at (4.15,-1.45) {$+$};
\pic[scale=0.70] at (4.35,-1.45) {stopwatch=green!45!black};
\draw[inner, draw=historycolor] (history.south) -- (strategy.north);
\draw[inner, draw=teal!70!black] (strategy.west) -- (llm.east);
\node[note, text=teal!50!black, anchor=north, inner sep=1pt] at (4.15,-3.32) {$R_{t+1}$};

\begin{scope}[on background layer]
\node[draw=black!30, dashed, fill=black!2, rounded corners=4pt, inner sep=4pt,
      fit=(eval) (history) (llm) (strategy)] (loopframe) {};
\end{scope}

\draw[flow, draw=historycolor] (6.40,-1.20) -- (7.20,-1.20);
\node[note, anchor=north] at (6.87,-1.31) {$\mathcal{D}_{\mathrm{val}}$};

\node[comptitle] at (7.70,-0.55) {Selected model $\hat{m}$};
\node[compbox, draw=outcolor, fill=outfill,
      minimum width=1.0cm, minimum height=2.9cm] (output) at (7.70,-2.225) {};
\pic at (7.70,-1.12) {stopwatch=green!45!black};
\node[note] at (7.70,-1.48) {$\mathcal{D}_{\mathrm{test}}$};
\draw[black!30, line width=0.5pt] (7.60,-3.46) rectangle (7.80,-1.72);
\fill[green!55!black!65] (7.60,-3.46) rectangle (7.80,-3.02);

\draw[dotted, teal!60!black, line width=1pt] (5.45,-4.1) -- (5.45,-4.45);
\node[compbox, draw=strategycolor, dashed, fill=teal!3, line width=0.7pt,
      minimum width=8.05cm, minimum height=1.50cm, anchor=north west]
      (pdrbox) at (0.15,-4.45) {};
\node[comptitle, anchor=west, text=teal!40!black] at (0.32,-4.68) {Profile-Diverse Retention (PDR)};
\node[note, anchor=west] at (3.95,-4.71) {(per-instance runtimes on $\mathcal{D}_{\mathrm{tr}}$)};
\draw[black!45, line width=0.5pt] (3.55,-5.45) -- (7.95,-5.45);
\foreach \gx/\ha/\hb/\hc/\lab in {3.65/0.42/0.38/0.40/$m_1$,
                                  4.45/0.14/0.40/0.20/$m_2$,
                                  5.25/0.30/0.16/0.35/$m_3$,
                                  6.70/0.16/0.18/0.17/$m_{t-1}$,
                                  7.50/0.40/0.36/0.42/$m_t$} {
  \fill[black!40] (\gx,-5.45) rectangle ++(0.09,\ha);
  \fill[black!40] (\gx+0.13,-5.45) rectangle ++(0.09,\hb);
  \fill[black!40] (\gx+0.26,-5.45) rectangle ++(0.09,\hc);
  \node[note, anchor=north] at (\gx+0.175,-5.47) {\lab};
}
\node[note] at (6.25,-5.28) {$\cdots$};
\draw[teal!60!black, rounded corners=1.5pt, line width=0.8pt]
  (3.6,-5.5) rectangle (4.05,-4.97);
\draw[teal!60!black, rounded corners=1.5pt, line width=0.8pt]
  (5.2,-5.5) rectangle (5.65,-4.97);
\draw[teal!60!black, rounded corners=1.5pt, line width=0.8pt]
  (6.65,-5.5) rectangle (7.1,-4.97);
\node[comptext, anchor=north, text width=2.75cm] at (1.8,-4.95)
  {$R_{t+1}$: the fastest attempt,\\ plus quality-diverse\\ ones (MMR selection)};

\end{tikzpicture}}
\caption{Iterative model reformulation. $B$: baseline; $\mathcal{D}_{\mathrm{tr}}$, $\mathcal{D}_{\mathrm{val}}$, $\mathcal{D}_{\mathrm{test}}$: training, validation, test sets.
At iteration $t\leq T$, the generator proposes model $m_t$ with rationale $\rho_t$, evaluates it on $\mathcal{D}_{\mathrm{tr}}$, and updates history $\mathcal{H}_t$ with evaluator feedback $E_t$; strategy $\sigma$ then selects context $R_{t+1}$ from $\mathcal{H}_t$ for the next iteration. The final model is chosen on $\mathcal{D}_{\mathrm{val}}$ after search terminates. \strategy{PDR} exemplifies the search strategies compared.}
\label{fig:overview}
\end{figure}
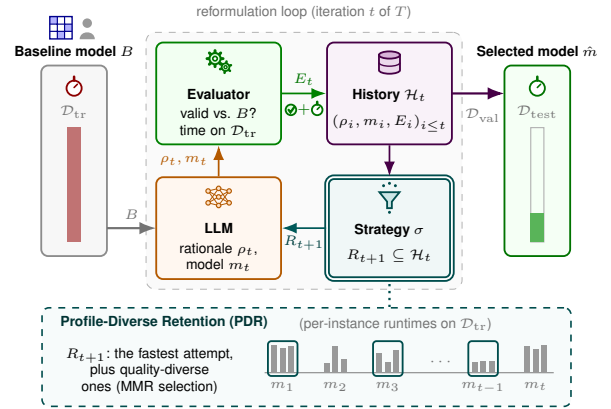

\paragraph{\Acp{llm} for Modelling.}
\Acp{llm} have renewed this goal by generating executable models directly from natural-language descriptions, across declarative paradigms including SAT~\citep{ye_satlm_2023}, ASP~\citep{ishay_leveraging_2023}, \ac{mip}~\citep{ahmaditeshnizi2024optimus,xiao2025survey}, and \ac{cp}~\citep{michailidis2024constraint,song2025llmcp,shi-etal-2025-constraintllm}.
Existing work and dedicated benchmarks mainly study whether the generated model is executable and produces correct answers, with recent results showing high correctness rates~\citep{michailidis2025cp,shi-etal-2025-constraintllm}.

\paragraph{Model Reformulation.}
A correct model can be slow to solve, while different formulations could result in better propagation and search behaviour~\citep{frisch2005rules}.
For example, expert modellers \emph{reformulate} models by changing the viewpoint (e.g., from integer to boolean variables), adding implied constraints, replacing decompositions with global constraints, breaking symmetries, and/or preprocessing instance data~\citep{DBLP:reference/fai/2}.
To reduce this manual effort, rule-based automation generates or transforms
models through predefined refinement and transformation
catalogues~\citep{nightingale2017automatically,akgun2022conjure,guns2019increasing}, while other
methods systematically construct portfolios of streamliners, which are specific constraints that reduce the search space of a model~\citep{spracklen2023automated}.
Most closely related to our work, \citet{voboril2025generating} use LLMs to generate and append streamliners to the initial model.
In contrast, we target whole-model reformulations, allowing the \ac{llm} to also change viewpoints, variables, or encodings rather than only append constraints.

\paragraph{\ac{ahd}.}
Our methodology is based on \ac{llm}-guided evolutionary search for imperative \ac{ahd}~\citep{liu2026systematic}: an \ac{llm} generates code, an automatic evaluator measures its quality, and selected attempts guide subsequent generations.
Existing approaches include program evolution~\citep{funsearch}, population-based operators~\citep{liuevolution}, reflective evolution~\citep{ye2024reevo}, single-parent search~\citep{van2024llamea}, or explicit crossover and diversity mechanisms~\citep{meyerson2024language,dat2025hsevo,eohs}.
These approaches differ mainly in which evaluated attempts inform the next generation and how the \ac{llm} is instructed to evolve them.
\acp{llm} have also been used to generate planning and SAT-solver heuristics, as well as \ac{mip} branching policies~\citep{gestrin2026llmevolved,sun2026autosat,hou2026llmbranch}.
While these works target imperative solving components, we use \acp{llm} for generating declarative CP models.

\paragraph{\Acp{llm} as Model Reformulators.}
We adapt \ac{ahd} to constraint model reformulation (Figure~\ref{fig:overview}): an \ac{llm} iteratively generates models, which are checked against the baseline and benchmarked on training instances.
For each instance we measure the time to first solution, which we verify using solution-level evaluation~\citep{michailidis2025cp}.
Then, the \emph{strategy} determines which evaluated attempts, instructions, and feedback enter the next prompt.
We compare strategies that span random sampling, recency, quality, full-history retention,
and operator-guided population search.
We also introduce \textit{\ac{pdr}}, a strategy that applies \ac{mmr} to instance-level runtime
vectors in order to retain attempts that are both fast and behaviourally
distinct~\citep{bradley2024quality,zhang2026rethinking}.
All speedups are reported on held-out test instances, and we propose final model selection on a separate validation set to avoid overfitting the training~data.

Our contributions are:
(i) an evolutionary framework for performance-oriented \ac{llm}-guided constraint model reformulation;
(ii) \ac{pdr}, a behavioural context-diversity strategy based on instance-level runtime vectors;
(iii) a systematic evaluation of ten strategies across eight CSPLib satisfaction problems, demonstrating large held-out speedups, the advantage of keeping the prompt context diverse, and the advantage of whole-model reformulation over an adapted streamliner generation baseline; and
(iv) that validation-based final model selection, a step usually absent from \ac{ahd}
pipelines, improves the held-out speedup of every strategy.

\section{Preliminaries}

\subsection{Constraint programming}
We use \ac{cp} as the formal representation of combinatorial satisfaction problems. A \ac{csp} is a tuple
$P = (\mathbf{X}, \mathbf{D}, \mathbf{C})$ where $\mathbf{X} = \{x_1, \ldots, x_n\}$ is a set of $n$ decision variables,  $\mathbf{D} = \{D_1, \ldots, D_n\}$ is a set of $n$ domains  where each $D_i$ is a finite set of allowed values for $x_i$, $\mathbf{C} = \{C_1, \ldots, C_m\}$ is a set of $m$ constraints where each $C_j$ specifies the allowed combinations of values for a subset of the variables.

An assignment $\mathbf{a} = \{x_1 = v_1, \ldots, x_n = v_n\}$ is a
\emph{solution} if $v_i \in D_i$ for all $i$ and every constraint in
$\mathbf{C}$ is satisfied; $\mathrm{Sol}(P)$
denotes the set of solutions, and the \ac{csp} is unsatisfiable when this set
is empty. Constraints may range from primitive arithmetic and logical relations
to \emph{global constraints} (e.g., \textsc{AllDifferent}) that capture recurring substructures and
come with dedicated propagation algorithms~\citep{beldiceanu2007global}.
In the model-and-solve workflow, the modeller states such a model in a
modelling framework (e.g. MiniZinc,~\citealp{minizinc} or CPMpy,~\citealp{guns2019increasing}) and a chosen solver (e.g. \ortools{},~\citealp{ortools}) is employed to search for solutions. The same problem can be modelled in many ways, varying in decision variables, constraints, and encodings.
These choices determine propagation strength and search behaviour, which impacts solving time~\cite{DBLP:reference/fai/2}.

\subsection{Large language models}
We use \acp{llm} as black-box text generators~\citep{vaswani_attention_2023}: given an input
prompt $p$, the model returns a token sequence
$\mathrm{LLM}(p) = (w_1, \ldots, w_k)$ sampled autoregressively, with
$w_j \sim P(w_j \mid p, w_{<j})$.
We prompt the \ac{llm} to return a structured answer containing: (i) a natural-language rationale; and (ii) an
executable model (Figure~\ref{fig:overview}).

\section{Problem Formulation}
\label{sec:problem-formulation}

Given an instance $d$, an executable model $m$ constructs a concrete \ac{csp} and
exposes an answer-extraction map:
\[
  \begin{aligned}
  m(d)&=(\mathbf{X}_{m,d},\mathbf{D}_{m,d},\mathbf{C}_{m,d}),\\
  \mathrm{Ans}_m(d)&=
  \{\eta_m(\mathbf{a}): \mathbf{a}\in \mathrm{Sol}(m(d))\}.
  \end{aligned}
\]
Here $\eta_m$ maps a solver-level assignment to the problem-level answer expected by the user.
This distinction is necessary as useful reformulations may introduce auxiliary variables, constraints, or encodings that are not part of the answer.
Two different model implementations can therefore use different decision variables while producing comparable answers.

Reformulation sometimes assumes the rewritten model preserves all solutions of the original~\citep{DBLP:reference/fai/2}.
However, verifying equisatisfiability requires comparing complete solution sets, which can be computationally challenging.
We instead assume a practical user-oriented setting, in which we apply solution-level validation on all instances~\citep{michailidis2025cp}.
\begin{definition}[Solution validity]
\label{def:solution-validity}
Let $B$ be the baseline model and $S$ a set of problem instances. Suppose that $m$ completes on every $d\in S$
and returns $\mathbf{y}_{m,d}\in\mathrm{Ans}_m(d)$. Then, $m$ is
\emph{solution-valid on $S$ with respect to $B$}, denoted as $\mathrm{SV}_S(m;B)$, if and only if: 
\(
  \forall d\in S:\quad \mathbf{y}_{m,d}\in\mathrm{Ans}_B(d).
\)
\end{definition}
In practice we add the reported answer as constraints on the baseline, by fixing the baseline decision
variables to the values assigned when solving $m$.
If $B(d)|_{\mathbf{y}_{m,d}}$ denotes the resulting model, then the answer is accepted exactly when $\mathrm{Sol}(B(d)|_{\mathbf{y}_{m,d}})\neq\varnothing$.
This metric therefore allows the candidate to use different viewpoints and encodings.

Let $\tau(m,d)$ be the solving time of model $m$ on instance $d$.
For an instance set $S$ we write $T_S(m)=\sum_{d\in S}\tau(m,d)$.
The goal is to produce a solution-valid model with minimized held-out total runtime:
\[
  m^\star \in
  \arg\min_{m\in\mathcal{M}}
  T_{\mathcal{D}_{\mathrm{test}}}(m)
  \quad\text{s.t.}\quad
  \mathrm{SV}_{\mathcal{D}_{\mathrm{test}}}(m;B),
\]
where $\mathcal{M}$ is the space of executable models satisfying the required input/output structure of the baseline model, and $\mathcal{D}_{\mathrm{test}}$ is a set of test instances unseen during a run.

\section{Iterative Model Reformulation}
\label{sec:methodology}

\begin{algorithm}[tb]
\caption{Iterative LLM-guided model reformulation.}
\label{alg:loop}
\begin{algorithmic}[1]
\REQUIRE Task instructions $\mathcal{F}$, baseline $B$, sets $\mathcal{D}_{\mathrm{tr}},\mathcal{D}_{\mathrm{val}}$, strategy $\sigma$, budget $T$.
\STATE Measure $\boldsymbol{\tau}_{B,\mathrm{tr}}$; set $\mathcal{H}\gets()$, $\mathcal{I}\gets\varnothing$, $\mathit{best}\gets\infty$.
\FOR{$t=1,\ldots,T$}
  \STATE $(R_t,o_t)\gets\sigma(\mathcal{H},t)$. \COMMENT{select the dynamic context}
  \STATE $P_t\gets\mathrm{Prompt}(\mathcal{F},B,\boldsymbol{\tau}_{B,\mathrm{tr}},R_t,o_t)$.
  \STATE $(\rho_t,m_t)\gets\mathrm{LLM}(P_t)$; $E_t\gets\mathrm{Eval}(m_t;B,\mathcal{D}_{\mathrm{tr}})$.
  \STATE $\mathcal{H}\gets\mathcal{H}\oplus(\rho_t,m_t,E_t)$.
  \IF{$\mathrm{SV}_{\mathcal{D}_{\mathrm{tr}}}(m_t;B)$ and $T_{\mathcal{D}_{\mathrm{tr}}}(m_t)<\mathit{best}$}
    \STATE $\mathcal{I}\gets\mathcal{I}\cup\{m_t\}$; $\mathit{best}\gets T_{\mathcal{D}_{\mathrm{tr}}}(m_t)$.
  \ENDIF
\ENDFOR
\STATE $\mathcal{I}\gets\{m\in\mathcal{I}:\mathrm{SV}_{\mathcal{D}_{\mathrm{val}}}(m;B)\}\cup\{B\}$.
\RETURN $\displaystyle\arg\min_{m\in\mathcal{I}}
T^{\mathrm{PAR2}}_{\mathcal{D}_{\mathrm{val}}}(m)$. \COMMENT{final model selection}
\end{algorithmic}
\end{algorithm}

Algorithm~\ref{alg:loop} summarises the iterative search loop.
At iteration $t$, the prompt\footnote{All prompts used are reported in the Appendix.} is
\(
  \mathrm{Prompt}(\mathcal{F},B,\boldsymbol{\tau}_{B,\mathrm{tr}},R_t,o_t)
\).
Here, $\mathcal{F}$ contains the task rules, the modelling-framework
documentation, and the problem description, $B$ is the baseline model, and
$\boldsymbol{\tau}_{B,\mathrm{tr}}$ contains its runtimes on the training instance set $\mathcal{D}_{\mathrm{tr}}$.
The history $\mathcal{H}_{t-1}=(h_1,\ldots,h_{t-1})$ contains all earlier
attempts, where $h_i=(\rho_i,m_i,E_i)$ records the rationale, parsed model,
and evaluation feedback (see below). The search strategy $\sigma$ maps this history and
the current iteration to an ordered sequence of retained attempts
$R_t\subseteq\mathcal{H}_{t-1}$ and a generation instruction $o_t$:
\(
  (R_t,o_t)=\sigma(\mathcal{H}_{t-1},t).
\)

The \ac{llm} returns a rationale $\rho_t$ and a candidate model $m_t$.
The evaluator $\mathrm{Eval}(m;B,S)$ executes $m$ on every instance in $S$, checks the returned
solution on each instance using Definition~\ref{def:solution-validity}, and measures the runtime under
a constant cap derived from the baseline (see
Section~\ref{sec:experimental-setup}).
The resulting record $E_t=\mathrm{Eval}(m_t;B,\mathcal{D}_{\mathrm{tr}})$ contains the validity label
and measured runtime for each training instance.
A retained attempt $h_i$ is shown in the prompt as its rationale $\rho_i$, its model $m_i$, and a
feedback block $f_i$ (derived from $E_i$) containing: (a) model validity outcome, (b) total runtime and speedup, and (c) summary of its per-instance behaviour.

A solution-valid attempt becomes a new incumbent when its total runtime is lower than all previous solution-valid attempts. 
After the run, all incumbents (including the baseline) are evaluated against a validation instance set $\mathcal{D}_{\mathrm{val}}$, and the best one is returned as the final model.
Timeout runs are penalised using PAR2~\cite{heule2019sat}, i.e., counted as twice the time limit.

\subsection{Search Strategies}

\begin{table}[t]
\centering
\small
\begin{tabularx}{\columnwidth}{@{}lY@{}}
\toprule
Strategy & Retained context ($R_t$) and operator \\
\midrule
\strategy{Sampling} & $R_t^{\mathrm{sam}}=\varnothing$ \\
\strategy{Last}-$k$ & $R_t^{\mathrm{rec}}(k)$ in Eq.~\eqref{eq:recency-retention} \\
\strategy{Top}-$k$ & $R_t^{\mathrm{qual}}(k)$, the $k$ fastest valid attempts \\
\strategy{Hybrid}-$(k,\ell)$ & $R_t^{\mathrm{hyb}}(k,\ell)$ in Eq.~\eqref{eq:hybrid-retention} \\
\strategy{Full-history} & $R_t^{\mathrm{full}}=\mathcal{H}_{t-1}$ \\
\strategy{EoH}\,[+feedback] & parents sampled by Eq.~\eqref{eq:eoh-parent-selection}, $o_t\in(e_1,e_2,m_1,m_2)$ \\
\strategy{PDR}-$k$\,[+warm-start] & up to $k$ valid attempts selected by Eq.~\eqref{eq:pdr-selection} \\
\bottomrule
\end{tabularx}
\caption{Summary of the search strategies. All strategies except
\strategy{EoH} use the base instruction $o_t=o_0$.}
\label{tab:strategies}
\end{table}

We first define two baseline strategies, followed by recency-based, quality-based, population-based,
and diversity-based strategies. 
All strategies except the population-based one use the default instruction $o_0$ which asks for one materially new, faster reformulation.
Table~\ref{tab:strategies} summarizes them.

\paragraph{Baseline strategies.}
\strategy{Sampling} uses no retained context and independently generates models from the same prompt $o_0$~\citep{wang2023selfconsistency}.
In contrast, \strategy{Full-history} retains every previous attempt that fits the context window:
  $R_t^{\mathrm{sam}}=\varnothing,
  \qquad
  R_t^{\mathrm{full}}=\mathcal{H}_{t-1}.$
If the full history exceeds the prompt budget, its oldest attempts are removed.

\paragraph{Recency and quality retention.}
Between the opposite baselines lie strategies that retain a small number of informative attempts, following the non-elitist $(1,1)$ and elitist $(1{+}1)$ AHD frameworks \citep{van2024llamea,funsearch}.
Recency retains the last $k$ attempts, so the \ac{llm} always sees its most recent feedback and can iteratively refine or repair its latest proposals:
\begin{equation}
  R_t^{\mathrm{rec}}(k)
  =(h_{\max\{1,t-k\}},\ldots,h_{t-1}).
  \label{eq:recency-retention}
\end{equation}
Quality retention, denoted $R_t^{\mathrm{qual}}(k)$, instead selects the $k$
attempts with lowest training runtime among the solution-valid attempts
$\mathcal{A}_{t-1}\subseteq\mathcal{H}_{t-1}$. The hybrid combines both:
\begin{equation}
  R_t^{\mathrm{hyb}}(k,\ell)
  =R_t^{\mathrm{rec}}(k)\cup R_t^{\mathrm{qual}}(\ell).
  \label{eq:hybrid-retention}
\end{equation}
All attempts are deduplicated and displayed chronologically.
Recency can retain rejected attempts, but quality retention does not.

\paragraph{Evolution of Heuristics (EoH).}
The previous strategies differ only in which attempts are retained, but they are all based on a single instruction $o_0$.
Another approach in \ac{ahd} is \ac{eoh} which adopts multiple variation instructions during the search and maintains an explicit population of models across iterations~\citep{liuevolution}.
We adapt \ac{eoh} to model reformulation as follows.

At iteration $t$, a population $\mathcal{P}_t$ of at most $K$ solution-valid models is maintained, ordered by training runtime.
The first $2K$ iterations use no parents and the base instruction $o_0$, so that the population is initialised.
The remaining iterations are grouped into batches of $K$.
Within a batch, every candidate is generated from the same population using one of the provided variation \emph{operators} (see below), and at the end of the batch the
new candidates and the current population are merged and truncated to the $K$ fastest
models~\citep{liu2024llm4ad}.
Each operator draws its parents from $\mathcal{P}_t$, with probability $\pi_t(m)$ inversely
proportional to total training runtime, so that faster models are drawn more often:
\begin{equation}
  \pi_t(m)
  \propto\frac{1}{T_{\mathcal{D}_{\mathrm{tr}}}(m)},
  \label{eq:eoh-parent-selection}
\end{equation}

The instruction $o_t$ is sequentially chosen from four operators $e_1,e_2,m_1,m_2$.
The first two operators use two parents:
$e_1$ asks for a reformulation that takes a fundamentally different modelling approach from both
parents, and $e_2$ asks for one that combines or extends parents' ideas.
The remaining two use a single parent:
$m_1$ asks for a local structural change that preserves the parent's main idea, and $m_2$ asks for a
change to one solver-facing component\footnote{We redefine $m_2$, as in the original formulation it mutates numerical
algorithm parameters.}.
\ac{eoh} originally presents parents without feedback, whereas the earlier strategies include it as a way to enrich context information.
We therefore study two variants, with and without feedback~$f_i$.

\paragraph{Profile-Diverse Retention (PDR).}
\Ac{eoh} maintains population diversity implicitly via its exploration-oriented operators ($e_1$ and $e_2$).
We additionally ask whether diversity can be controlled explicitly through the retained context, following behaviour-aware and quality-diversity search~\citep{zhang2026rethinking,bradley2024quality}.
\Ac{pdr} describes each attempt by its behaviour across the training instances, and retains attempts that are fast but behave differently from one another.
Let $\mathcal{A}_t$ be the solution-valid attempts at iteration $t$ (Definition~\ref{def:solution-validity}).
The instance-level profile of model $m$ is defined as
\[
p_{t,d}(m)
= \frac{\log\tau_d(m)-\min_{u \in \mathcal{A}_t}\log\tau_d(u)}
{\max_{u \in \mathcal{A}_t}\log\tau_d(u)-\min_{u \in \mathcal{A}_t}\log\tau_d(u)},
\]
where $\tau_d(m)$ is the training runtime of model $m$ on instance $d$.
Solving times may differ by orders of magnitude, so we use logarithms and normalise each instance separately.
Without the logarithm a single slow attempt flattens the differences among the fast ones, and without the per-instance normalisation the hardest instances dominate every comparison.
If all attempts in $\mathcal{A}_t$ have the same runtime on an instance, $p_{t,d}(m)$ is set to $0$.
We compare two attempts by the root-mean-square distance between their profiles:
\begin{equation}
  d_t(m,u)
  =\sqrt{\frac{1}{|\mathcal{D}_{\mathrm{tr}}|}
  \sum_{d\in\mathcal{D}_{\mathrm{tr}}}
  \bigl(p_{t,d}(m)-p_{t,d}(u)\bigr)^{2}}
  \label{eq:pdr-distance}
\end{equation}

Quality uses the same logarithmic scale, so that it keeps discriminating among the fast attempts.
With $\ell_t(m)=\log T_{\mathcal{D}_{\mathrm{tr}}}(m)$, we set
$q_t(m)=1-(\ell_t(m)-\ell_t^{\min})/(\ell_t^{\max}-\ell_t^{\min})$, so larger values denote faster models.
\Ac{pdr} first retains the fastest attempt, and then repeatedly adds
\begin{equation}
  \arg\max_{m\in\mathcal{A}_t\setminus R_t}
  \left[\lambda q_t(m)+(1-\lambda)\min_{u\in R_t}d_t(m,u)\right],
  \label{eq:pdr-selection}
\end{equation}
where $R_t$ is the set retained so far, until $|R_t|=k$, and
$\lambda\in[0,1]$ controls the trade-off between quality and diversity.
This adapts MMR reranking to quality-diverse model retention,
where quality is total runtime and diversity is instance-level runtime
behaviour~\citep{mmr}. The bottom of Figure~\ref{fig:overview} visualises the selection step. 
Finally, \ac{eoh} builds its initial population from models generated without any retained context.
We investigate an analogous \strategy{PDR warm-start}, which retains no context for the first $2k$ iterations and applies \ac{pdr} afterwards.

\begin{figure*}[!t]
\centering
\includegraphics[width=0.8\textwidth]{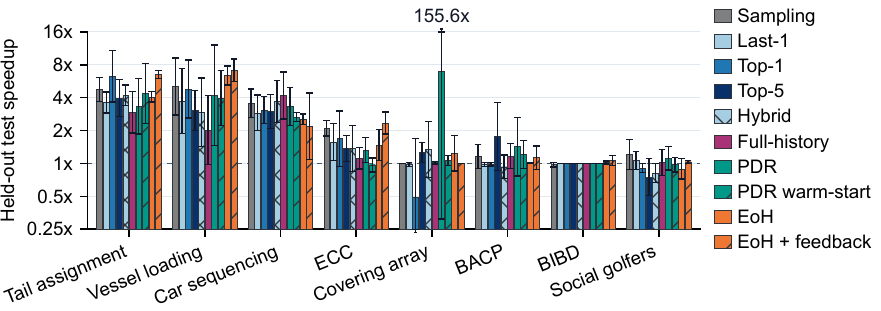}
\caption{Held-out test speedup by problem and strategy, ordered by each
problem's GM across all runs. Each bar is the GM over the three
repetitions, and the interval shows the geometric standard deviation.}
\label{fig:all10-panels}
\end{figure*}

\section{Experiments}
\label{sec:experimental-setup}

We organise the evaluation around three questions:
\begin{enumerate}
  \item \textbf{RQ1:} Does \ac{llm}-guided reformulation produce valid and faster models than the baseline on held-out instances?
  \item \textbf{RQ2:} How does the search strategy, e.g. what is retained, how
  it is selected, and whether operators guide the generation, affect the speedup and its cost?
  \item \textbf{RQ3:} Does validation-based model selection avoid overfitting on the training instances?
\end{enumerate}

We study eight satisfaction problems from CSPLib~\citep{csplib}, including \emph{tail assignment} (problem 115),
\emph{vessel loading} (problem 8), \emph{car sequencing} (problem 1), \emph{error-correction codes} (problem 36), \emph{covering arrays} (problem 45), \emph{balanced academic
curriculum} (problem 30), \emph{balanced incomplete block design} (problem 28), and \emph{social
golfer} (problem 10). Following prior work~\citep{spracklen2023automated}, we select these benchmarks because they span a broad range of practical domains, including combinatorial design, scheduling, manufacturing, timetabling, transportation, and coding problems. 
Each problem is paired with a parameterised baseline model,
and generated models are instructed to use the same input and answer-extraction structure.
All models are solved with a single-core \ortools{} configuration (one search worker)
through \cpmpy{} 0.10.1~\citep{guns2019increasing,ortools}. Up to 8 runs were
executed concurrently, where each solve remains single-core. All experiments
ran on Intel Core i7-12700 (12 cores, 64\,GB RAM), Ubuntu 22.04.
The LLM used is DeepSeek V4 Flash\footnote{Accessed through the
DeepSeek API, 8--12 July 2026.}, with high reasoning effort, a 1M-token
context and a 64k-token output limit~\cite{xu2026deepseek}.
We leave \textit{temperature} and \textit{top\_p} at their API defaults of $1.0$.~\footnote{Code and detailed results will be made publicly available.}

\paragraph{Configurations.}
We evaluate ten configurations: \strategy{Sampling}, \strategy{Full-history},
the retention strategies \strategy{Last-1}, \strategy{Top-1},
\strategy{Top-5}, and \strategy{Hybrid}, \strategy{EoH} with and without feedback, and \strategy{PDR} with and without warm-start.
\strategy{Last-1}, \strategy{Top-1}, and \strategy{Hybrid} use $k=1$.
\strategy{EoH} uses $K=5$, and \strategy{PDR} uses $k=5$. \strategy{Top-5} provides a
quality-only comparison to \strategy{PDR}, and \strategy{PDR} uses $\lambda=0.5$ to weight quality
and diversity equally.
During a run, each candidate's total training evaluation is bounded with $C_B=\max(T_B,\min(1.5T_B,900))$ seconds.
Each configuration runs for 100 iterations on every problem, with three repetitions:
$8\times10\times3=240$ distinct runs.

\paragraph{Instances and Measures.}
We use instance pools generated for automated streamliner research
with the AutoIG framework~\citep{spracklen2023automated,dang2022framework}.
As difficulty depends on the model and solver, we remeasure every instance
with the CPMpy/OR-Tools baseline. For each problem, we choose satisfiable
instances with baseline runtimes in $[1,300]$ seconds,
partition them into eight equal-frequency runtime bins, and sample 1
training, 1 validation, and 2 test instances per bin: an 8/8/16 split per problem.
For a solution-valid model, baseline-relative speedup $s$ on set $S$ is
$T_S(B)/T_S(m)$.
We aggregate $n$ speedups by their geometric mean,
$\mathrm{GM}=\left(\prod_{i=1}^{n}s_i\right)^{1/n}$.
A timeout on an instance results in twice the time limit (PAR2) during validation and test evaluation.

\section{Results \& Discussion}
\label{sec:results}

Every run returns one model $\widehat m$ (Algorithm~\ref{alg:loop}).
Table~\ref{tab:all10-main} gives the per-strategy geometric mean (GM) over all runs, Figure~\ref{fig:all10-panels} gives the
per-problem strategy results. We compute 95\%
confidence intervals (CIs) with 10,000 problem-level bootstrap resamples,
pairing the three repetitions of each sampled problem.

\begin{table}
\centering
\resizebox{\linewidth}{!}{%
\begin{tabular}{lrrrr}
\toprule
Strategy & \shortstack{GM speedup\\{}[95\% CI]} & Median & \shortstack{Faster than\\baseline} & \shortstack{Leave-one-\\problem-out} \\
\midrule
\strategy{Sampling} & 1.99$\times$ [1.27, 3.18] & \textbf{1.75$\times$} & 14/24 & 1.74--2.20$\times$ \\
\strategy{Last-1} & 1.68$\times$ [1.16, 2.52] & 1.16$\times$ & 12/24 & 1.49--1.81$\times$ \\
\strategy{Top-1} & 1.69$\times$ [0.95, 3.07] & 1.01$\times$ & 12/24 & 1.41--2.03$\times$ \\
\strategy{Top-5} & 1.74$\times$ [1.18, 2.55] & 1.44$\times$ & \textbf{17/24} & 1.54--1.96$\times$ \\
\strategy{Hybrid-(1,1)} & 1.68$\times$ [1.12, 2.59] & 1.19$\times$ & 13/24 & 1.48--1.86$\times$ \\
\strategy{Full-history} & 1.55$\times$ [1.12, 2.28] & 1.04$\times$ & 14/24 & 1.35--1.65$\times$ \\
\strategy{PDR} & \textbf{2.26$\times$ [1.44, 3.65]} & 1.48$\times$ & 16/24 & 1.93--2.54$\times$ \\
\strategy{PDR warm-start} & 1.65$\times$ [1.13, 2.58] & 1.12$\times$ & 15/24 & 1.44--1.78$\times$ \\
\strategy{EoH} & 1.80$\times$ [1.16, 2.98] & 1.22$\times$ & 16/24 & 1.50--1.99$\times$ \\
\strategy{EoH + feedback} & 2.04$\times$ [1.26, 3.62] & 1.35$\times$ & 15/24 & 1.70--2.25$\times$ \\
\bottomrule
\end{tabular}
}

\caption{Held-out test results per strategy over 24 runs each. 
For the final column, we omit each problem in turn and report the range of the eight recomputed GMs to assess sensitivity to individual problems. All strategies retain a GM speedup above $1\times$.
}
\label{tab:all10-main}
\end{table}

\begin{table*}[!t]
\centering
\scriptsize
\setlength{\tabcolsep}{3pt}
\begin{tabularx}{\textwidth}{@{}p{0.105\textwidth}>{\raggedright\arraybackslash}p{0.28\textwidth}>{\raggedright\arraybackslash}p{0.34\textwidth}c>{\raggedright\arraybackslash}X@{}}
\toprule
Problem &
Most frequent pattern (A) &
Second recurring pattern (B) &
\shortstack{Faster\\models} &
\shortstack{Best model\\(strategy, patterns)} \\
\midrule

\emph{Tail assignment} &
Boolean variables over feasible flight arcs (15/30) &
Set-membership constraints on successor variables (12/30) &
30/30 & $11.5\times$ (\strategy{Top-1}, A) \\[1pt]

\emph{Vessel loading} &
Reified pairwise spatial disjunctions~\citep{moffitt2006optimal} (26/29) &
Binary orientation variables for $90^\circ$ rotation~\citep{andrade2013symmetry} (26/29) &
29/30 &
$11.8\times$ (\strategy{PDR}, A+B) \\[1pt]

\emph{Car sequencing} &
Boolean class--position matrix viewpoint~\citep{artigues2014sat} (29/29) &
Aggregation of interchangeable types by option profile~\citep{dincbas1988solving} (6/29) &
29/30 &
$5.8\times$ (\strategy{Full-history}, A+B) \\[1pt]

\emph{ECC} &
Symmetry breaking via lexicographic ordering of codewords%
~\citep{flener2002breakingsym} (15/24) &
Tabulation of pairwise symbol-distance contributions%
~\citep{ijcai2023p211} (12/24) &
24/30 & $3.0\times$ (\strategy{EoH+feedback}, B) \\[1pt]

\emph{Covering array} &
Initial-column symmetry breaking (7/8) &
Base-$g$ enc. of column-wise $t$-tuples%
  ~\citep{hnich2006constraint} (6/8) &
8/30 & $250.7\times$ (\strategy{PDR}, A+B) \\[1pt]

\emph{BACP} &
Prerequisite-derived bounds on course-period domains%
~\citep{schaus2008global} (11/13) &
Redundant cumulative constraints for period capacities%
~\citep{aggoun1993extending} (2/13) &
13/30 & $3.9\times$ (\strategy{Top-5}, A) \\[1pt]

\emph{BIBD} &
Equivalent incidence-matrix formulations (2/2) &
-- &
2/30 & $1.2\times$ (\strategy{EoH+feedback}, A) \\[1pt]

\emph{Social golfers} &
Sym. breaking via first-week ordering or fixed assignments~\citep{liu2019solving} (9/9) &
Global-cardinality constraints for group sizes~\citep{regin1996generalized} (4/9) &
9/30 & $1.7\times$ (\strategy{Sampling}, A) \\

\bottomrule
\end{tabularx}

\caption{Recurring modelling patterns in validation-selected models that
outperform the baseline on the test set. Pattern counts report occurrences
within these models and citations identify established uses of the corresponding
modelling technique.}
\label{tab:all10-winners}
\end{table*}

\subsection{RQ1: Effectiveness and Reliability}
\label{sec:rq1}

\textit{Evolutionary \ac{llm}-guided search finds solution-valid models with large held-out speedups.}
Across the 240 runs, 144 return a faster model on the held-out instances. The
rest return the baseline (67) or a model that validated faster but did not
improve on test (29).
Notably, none of the selected models returned an invalid answer or a false-\textsc{unsat}
outcome on test instances. Six timed out on at least one test instance.
Overall, every strategy improves on the baseline, from $1.55\times$ (\strategy{Full-history}) to $2.26\times$ (\strategy{PDR}), and every confidence interval except \strategy{Top-1}'s lies above $1.0\times$ (Table~\ref{tab:all10-main}).

\paragraph{Results per problem.}
Figure~\ref{fig:all10-panels} separates the eight problems.
On \emph{tail assignment}, \emph{vessel loading}, and \emph{car sequencing}, nearly every
run beats the baseline (30, 29, and 29 of 30), and the per-problem GM
over all strategies reaches $4.3\times$, $4.1\times$, and $3.1\times$.
On \emph{error-correction codes} the gains are more moderate (24 of 30,
$1.5\times$). On \emph{BACP} and \emph{covering arrays} most runs
return the baseline or a small gain, but individual runs find large
improvements, e.g. one \strategy{PDR} run reaches $250.7\times$ on covering
arrays.
Lastly, very few runs improve on the baseline for \emph{social golfers} (9 of 30) and \emph{BIBD} (2 of 30).

\paragraph{Successful reformulations.}
We inspected every returned model that is faster than the baseline on the test instances and catalogued its modelling techniques (Table~\ref{tab:all10-winners}).
Most of them use techniques that are established in the \ac{cp} literature.
For example, some \emph{error-correction} models replace repeated distance expressions with table constraints, a known automatic-tabulation technique~\citep{ijcai2023p211}.
Across \emph{car sequencing}, \emph{vessel loading}, \emph{BACP}, and \emph{tail assignment}, the faster models use alternative variable viewpoints, Boolean encodings, or redundant constraints to strengthen propagation.
The largest gain is on \emph{covering arrays}, where the $250.7\times$ model fixes a canonical enumeration prefix that restricts the solutions it can return, acting as a streamliner~\citep{gomes2004streamlined}.

\paragraph{Failures.}
On \emph{BIBD}, 76\% of
all candidate evaluations time out on single training instances, and only 19
of 3,000 proposals complete as solution-valid, so most runs never obtain
an incumbent. 79\% of the proposals add lexicographic ordering constraints
across the columns of the incidence matrix. On these instances the matrix is
very wide (for example, $8\times2744$), so the added constraints span
thousands of columns, and the added propagation work may offset any reduction in search.
On \emph{Covering arrays}, the small training instances (median baseline total of 24 seconds) results in frequent model timeouts on the larger validation instances, causing the run to return the baseline as a fallback.

\paragraph{Comparing with additive streamlining.}
An alternative way to improve solving time is to append individual constraints rather than rewrite the entire model.
In this direction, StreamLLM generates \emph{streamliners}, additional constraints that restrict
the solution space and may accelerate search~\citep{voboril2025generating}.
StreamLLM selects the three best streamliners during training, and then races their corresponding models with the original on each test instance (portfolio).
For a fair non-portfolio comparison, we retain the original search procedure but adapt the final evaluation to our setting (\emph{StreamLLM-single}): each run independently keeps training incumbents, which are validated together with the baseline.
Then, the model with the lowest total validation runtime is evaluated on the test instances.

In Table~\ref{tab:all10-streamllm}, we compare StreamLLM-single with the two highest-ranked reformulation strategies, \strategy{PDR} and \strategy{EoH+feedback}.
Their aggregate GMs are $2.26\times$ and $2.04\times$, respectively, compared with $1.02\times$ for StreamLLM-single.
\ac{pdr} achieves higher speedups than StreamLLM-single on all eight problems and in 19 of 24 repetitions, with the two methods nearly tied on \emph{error-correction codes}, and \strategy{EoH+feedback} achieves a higher problem-level GM on six of eight problems.
StreamLLM-single still improves over the baseline on five problem-level GMs, but times out on one test instance in four occasions: one \emph{social-golfers} and all three \emph{BIBD} repetitions.
More details are available in the Appendix.

\begin{table}
\centering
\scriptsize
\setlength{\tabcolsep}{2.2pt}
\setlength{\tabcolsep}{2.8pt}
\resizebox{\linewidth}{!}{%
\begin{tabular}{@{}l cccccccc @{\hspace{5pt}} c@{}}
\toprule
Method & Car & Vessel & Golf & BIBD & BACP & ECC & CA & Tail & All \\
\midrule
StreamLLM-single & 1.08 & 0.96 & 0.76 & 0.50 & 1.23 & 1.32 & 1.25 & 1.44 & 1.02 \\
\dashedmidrule
EoH + feedback & 2.19 & 7.15 & 1.03 & 1.06 & 1.13 & 2.34 & 1.00 & 6.51 & 2.04 \\
\strategy{PDR} & 3.33 & 4.21 & 1.12 & 1.00 & 1.43 & 1.33 & 6.96 & 3.34 & \textbf{2.26} \\
\bottomrule
\end{tabular}
}

\caption{Comparison of effective PAR2 test speedup (higher is better) GM over three repetitions.}
\label{tab:all10-streamllm}
\end{table}

\subsection{RQ2: Search Strategies}
\label{sec:rq2}

\textit{Strategies that ensure the retained context is diverse, outperform those that retain only the most recent or the fastest attempts.}
Figure~\ref{fig:all10-panels} shows that performance varies more across problems than across strategies. \strategy{PDR} achieves the highest aggregate GM, but is largely driven by \textit{covering arrays}.
\strategy{EoH+feedback} ranks second overall and is best on four problems, whereas \strategy{Sampling} ranks third but has the highest median.
A common feature of these strategies is that they promote diversity through different mechanisms.
\strategy{Sampling} relies on stochastic generation from a fixed prompt, \strategy{EoH+feedback} uses exploration-oriented operators, and \strategy{PDR} retains attempts with different solving behaviour.

\begin{figure}[!t]
\centering
\includegraphics[width=0.7\columnwidth]{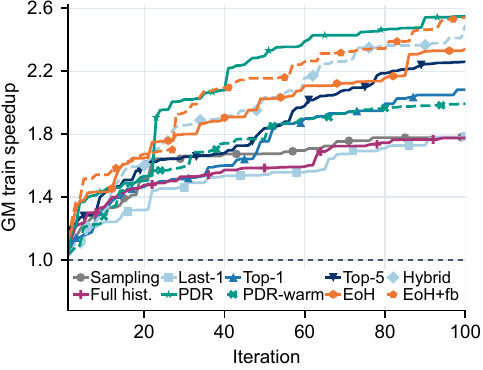}
\caption{GM training speedup of the current incumbent over iterations, per strategy.}
\label{fig:all10-dynamics}
\end{figure}

\paragraph{Search progression.}
Although not all other strategies provide a consistent advantage over sampling, they show different training progression (Figure~\ref{fig:all10-dynamics}): \strategy{Sampling} finds its final incumbent early (median iteration 26), while every other strategy keeps improving into iterations 48--78.
In 43 runs the last improvement came at iteration 90 or later, so a larger budget could help some of them further.

\paragraph{Ablations.} We compare closely related configurations to examine the effects of individual choices.
\strategy{PDR} improves by $1.30\times$ over the quality-only
\strategy{Top-5}, showing the benefit of retaining a diverse context, and by $1.37\times$ over
\strategy{PDR warm-start}, suggesting that an initial context-free warm-start (diversity through randomness)
is less beneficial.
Additionally, for the EoH strategy we see that performance feedback is beneficial across problems: \strategy{EoH+feedback} improves over \strategy{EoH}, with the largest gains on \emph{error-correction codes} and \emph{tail assignment} (both $1.60\times$).
The remaining comparisons perform closely: quality versus recency (\strategy{Top-1} vs \strategy{Last-1}), their combination (\strategy{Hybrid}), and the retained context size
(\strategy{Top-5} vs.\ \strategy{Top-1}) all have GM ratios near $1.0\times$.

\begin{figure}
\centering
\includegraphics[width=0.65\columnwidth]{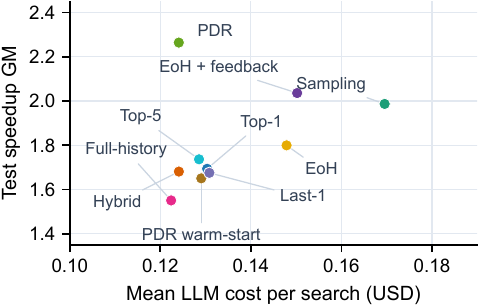}
\caption{Mean \ac{llm} cost per run against speedup.}
\label{fig:all10-cost}
\end{figure}

\paragraph{Cost.}
Figure~\ref{fig:all10-cost} relates cost to speedup.
\strategy{Sampling} is the most expensive strategy (\$0.17 per run)
because it generates the most output: 0.60M tokens per run, of which 0.50M
are reasoning tokens, about twice \strategy{Full-history}'s 0.29M. Starting
again from the same prompt appears to make each sample re-derive decisions that
retained-history strategies can reuse. Input is inexpensive due to shared-prefix KV caching used in recent LLMs~\citep{gim2024prompt,xu2026deepseek}.
\strategy{Full-history}
therefore stays inexpensive even though it sends 6.7M input tokens per run,
ten times more than any other strategy, of which 98\% are cache hits.
Notably, \strategy{PDR} delivers the highest GM at 27\% lower cost than \strategy{Sampling}.

\subsection{RQ3: Model Selection}
\label{sec:rq3}

\begin{figure}
\centering
\includegraphics[width=0.7\columnwidth]{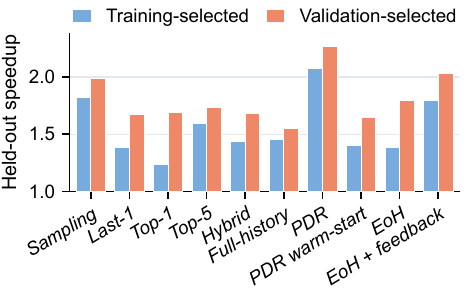}
\caption{Held-out speedup when selecting the final model by validation or
keeping the final training incumbent.}
\label{fig:all10-selection}
\end{figure}

\textit{Validation-based model selection consistently improves held-out performance and mitigates overfitting.}
A common approach in related evolutionary \ac{llm} search is to immediately return the fastest training incumbent.
FunSearch exposes the highest-scoring program found, while \strategy{EoH} evaluates the
best heuristic in its final population~\citep{funsearch,liuevolution}.
For every run, we evaluate both the final training incumbent and the model
returned by Algorithm~\ref{alg:loop}, which selects on validation with the
baseline as a fallback. Validation improves the aggregate test GM of every
strategy (Figure~\ref{fig:all10-selection}).
Additionally, when no incumbent validates faster than the baseline, the run returns the baseline, which produced the 67 neutral outcomes and discarded the \emph{covering-array} and \emph{BIBD} candidates that had only fit their training instances.

\section{Conclusion}

We presented an evolutionary framework for reformulating constraint models and
proposed \ac{pdr} as well as \ac{eoh} variants to retain models with diverse instance-level runtime
vectors. We compared ten search configurations across eight CSPLib problems,
evaluating models selected on held-out validation instances.
The results show that reformulation improves solver performance overall, and qualitative analysis additionally revealed the usage of established modelling techniques such as viewpoint changes, tabulation, symmetry breaking, and precomputation.
Among the strategies, those that make the retained context diverse, outperform those that retain only the most recent or the fastest attempts, while performance feedback improves operator-guided search.
Individual strategies remain hard to separate, and the highest aggregate speedup, obtained by \ac{pdr}, is determined by a small number of runs with very large speedups.
Validation-based selection further improves the speedup of every strategy and filters candidates that overfit on the training instances.
Moreover, we adapt a streamliner-generation baseline to our experimental setting, and show that whole-model reformulation outperforms merely adding streamlining constraints on all problems.

\paragraph{Limitations \& Future Work.}\label{sec:limitations}
We study one \ac{llm}, one solver configuration, and eight satisfaction problems
under fixed instances, hardware, and search budget.
Future work can test other models, solvers (with varying configurations), agentic strategies, and optimisation problems. Search strategies could also use prior feedback more explicitly, keep a compressed history/summary, or deliberately contrast strong incumbents with labelled failed or slow attempts.
Moreover, \Ac{pdr} measures diversity only through instance runtimes, and incorporating solver-rich information such as propagations, conflicts, or search-tree depth could provide even more behavioural signals.
\section*{Acknowledgments}
This project has received funding from the European Research Council (ERC)
under the European Union's Horizon 2020 research and innovation program
(Grant No.\ 101002802, CHAT-Opt), and from the Flemish Government under
``Onderzoeksprogramma Artifici\"ele Intelligentie (AI) Vlaanderen''.

\FloatBarrier
\bibliography{references}

@article{aggoun1993extending,
  title={Extending CHIP in order to solve complex scheduling and placement problems},
  author={Aggoun, Abderrahmane and Beldiceanu, Nicolas},
  journal={Mathematical and computer modelling},
  volume={17},
  number={7},
  pages={57--73},
  year={1993},
  publisher={Elsevier}
}

@inproceedings{ahmaditeshnizi2024optimus,
  title={OptiMUS: Scalable Optimization Modeling with (MI) LP Solvers and Large Language Models},
  author={Ahmaditeshnizi, Ali and Gao, Wenzhi and Udell, Madeleine},
  booktitle={International Conference on Machine Learning},
  pages={577--596},
  year={2024},
  organization={PMLR}
}

@article{akgun2022conjure,
  title={Conjure: Automatic generation of constraint models from problem specifications},
  author={Akg{\"u}n, {\"O}zg{\"u}r and Frisch, Alan M and Gent, Ian P and Jefferson, Christopher and Miguel, Ian and Nightingale, Peter},
  journal={Artificial Intelligence},
  volume={310},
  pages={103751},
  year={2022},
  publisher={Elsevier}
}

@article{andrade2013symmetry,
  title={Symmetry-breaking constraints for packing identical rectangles within polyhedra},
  author={Andrade, Ricardo and Birgin, Ernesto G},
  journal={Optimization Letters},
  volume={7},
  number={2},
  pages={375--405},
  year={2013},
  publisher={Springer}
}

@inproceedings{artigues2014sat,
  title={SAT and hybrid models of the car sequencing problem},
  author={Artigues, Christian and Hebrard, Emmanuel and Mayer-Eichberger, Valentin and Siala, Mohamed and Walsh, Toby},
  booktitle={International Conference on Integration of Constraint Programming, Artificial Intelligence, and Operations Research},
  pages={268--283},
  year={2014},
  organization={Springer}
}

@article{beldiceanu2007global,
  title={Global constraint catalogue: Past, present and future},
  author={Beldiceanu, Nicolas and Carlsson, Mats and Demassey, Sophie and Petit, Thierry},
  journal={Constraints},
  volume={12},
  pages={21--62},
  year={2007},
  publisher={Springer}
}

@inproceedings{bradley2024quality,
  title={Quality-diversity through AI feedback},
  author={Bradley, Herbie and Dai, Andrew and Teufel, Hannah and Zhang, Jenny and Oostermeijer, Koen and Bellagente, Marco and Clune, Jeff and Stanley, Kenneth and Schott, Gr{\'e}gory and Lehman, Joel},
  booktitle={International Conference on Learning Representations},
  volume={2024},
  pages={21036--21147},
  year={2024}
}

@InProceedings{csplib,
author="Gent, Ian P.
and Walsh, Toby",
editor="Jaffar, Joxan",
title="CSPlib: A Benchmark Library for Constraints",
booktitle="Principles and Practice of Constraint Programming -- CP'99",
year="1999",
publisher="Springer Berlin Heidelberg",
address="Berlin, Heidelberg",
pages="480--481",
isbn="978-3-540-48085-3"
}

@inproceedings{dang2022framework,
  title={A Framework for Generating Informative Benchmark Instances},
  author={Dang, Nguyen and Akg{\"u}n, {\"O}zg{\"u}r and Espasa, Joan and Miguel, Ian and Nightingale, Peter},
  booktitle={28th International Conference on Principles and Practice of Constraint Programming (CP 2022)},
  pages={18--1},
  year={2022},
  organization={Schloss Dagstuhl--Leibniz-Zentrum f{\"u}r Informatik}
}

@inproceedings{dat2025hsevo,
  title={Hsevo: Elevating automatic heuristic design with diversity-driven harmony search and genetic algorithm using {LLMs}},
  author={Dat, Pham Vu Tuan and Doan, Long and Binh, Huynh Thi Thanh},
  booktitle={Proceedings of the AAAI Conference on Artificial Intelligence},
  volume={39},
  pages={26931--26938},
  year={2025}
}

@book{DBLP:reference/fai/2,
  editor       = {Francesca Rossi and
                  Peter van Beek and
                  Toby Walsh},
  title        = {Handbook of Constraint Programming},
  series       = {Foundations of Artificial Intelligence},
  volume       = {2},
  publisher    = {Elsevier},
  year         = {2006},
  url          = {https://www.sciencedirect.com/science/bookseries/15746526/2},
  isbn         = {978-0-444-52726-4},
  bibsource    = {dblp computer science bibliography, https://dblp.org}
}

@inproceedings{dincbas1988solving,
  title={Solving the Car-Sequencing Problem in Constraint Logic Programming.},
  author={Dincbas, Mehmet and Simonis, Helmut and Van Hentenryck, Pascal},
  booktitle={ECAI},
  volume={88},
  pages={290--295},
  year={1988}
}

@inproceedings{eohs,
author = {Liu, Fei and Liu, Yilu and Zhang, Qingfu and Tong, Xialiang and Yuan, Mingxuan},
title = {EoH-S: evolution of heuristic set using LLMs for automated heuristic design},
year = {2026},
isbn = {978-1-57735-906-7},
publisher = {AAAI Press},
url = {https://doi.org/10.1609/aaai.v40i43.41038},
doi = {10.1609/aaai.v40i43.41038},
booktitle = {Proceedings of the Fortieth AAAI Conference on Artificial Intelligence and Thirty-Eighth Conference on Innovative Applications of Artificial Intelligence and Sixteenth Symposium on Educational Advances in Artificial Intelligence},
articleno = {4137},
numpages = {9},
}

@InProceedings{flener2002breakingsym,
author="Flener, Pierre
and Frisch, Alan M.
and Hnich, Brahim
and Kiziltan, Zeynep
and Miguel, Ian
and Pearson, Justin
and Walsh, Toby",
editor="Van Hentenryck, Pascal",
title="Breaking Row and Column Symmetries in Matrix Models",
booktitle="Principles and Practice of Constraint Programming - CP 2002",
year="2002",
publisher="Springer Berlin Heidelberg",
address="Berlin, Heidelberg",
pages="462--477",
isbn="978-3-540-46135-7"
}

@article{freuder1997holy_grail,
author = {Freuder, Eugene C.},
title = {In Pursuit of the Holy Grail},
year = {1997},
issue_date = {April 1997},
publisher = {Kluwer Academic Publishers},
address = {USA},
volume = {2},
number = {1},
issn = {1383-7133},
url = {https://doi.org/10.1023/A:1009749006768},
doi = {10.1023/A:1009749006768},
journal = {Constraints},
month = {apr},
pages = {57–61},
numpages = {5}
}

@article{freuder2018progress,
  title = {Progress towards the Holy Grail},
  author = {Eugene C. Freuder},
  journal = {Constraints An Int. J.},
  volume = {23},
  number = {2},
  pages = {158--171},
  year = {2018},
  publisher = {Springer},
  doi = {10.1007/s10601-017-9275-0}
}

@inproceedings{frisch2005rules,
  title={The rules of constraint modelling},
  author={Frisch, Alan M and Jefferson, Christopher and Hern{\'a}ndez, Bernadette Mart{\'\i}nez and Miguel, Ian},
  booktitle={IJCAI},
  pages={109--116},
  year={2005}
}

@article{funsearch,
  title={Mathematical discoveries from program search with large language models},
  author={Romera-Paredes, Bernardino and Barekatain, Mohammadamin and Novikov, Alexander and Balog, Matej and Kumar, M Pawan and Dupont, Emilien and Ruiz, Francisco JR and Ellenberg, Jordan S and Wang, Pengming and Fawzi, Omar and others},
  journal={Nature},
  volume={625},
  number={7995},
  pages={468--475},
  year={2024},
  publisher={Nature Publishing Group UK London}
}

@inproceedings{
gestrin2026llmevolved,
title={{LLM}-Evolved Domain-Independent Heuristics for Symbolic {AI} Planning},
author={Elliot Gestrin and Jendrik Seipp},
booktitle={Workshop on Planning in the Era of LLMs at ICAPS-26},
year={2026},
url={https://openreview.net/forum?id=6qYhouFPu8}
}

@article{gim2024prompt,
  title={Prompt cache: Modular attention reuse for low-latency inference},
  author={Gim, In and Chen, Guojun and Lee, Seung-seob and Sarda, Nikhil and Khandelwal, Anurag and Zhong, Lin},
  journal={Proceedings of Machine Learning and Systems},
  volume={6},
  pages={325--338},
  year={2024}
}

@inproceedings{gomes2004streamlined,
  title={Streamlined constraint reasoning},
  author={Gomes, Carla and Sellmann, Meinolf},
  booktitle={International Conference on Principles and Practice of Constraint Programming},
  pages={274--289},
  year={2004},
  organization={Springer}
}

@inproceedings{guns2019increasing,
  title = {Increasing modeling language convenience with a universal n-dimensional array, CPpy as python-embedded example},
  author = {Guns, Tias},
  booktitle = {Proceedings of the 18th workshop on Constraint Modelling and Reformulation at CP (Modref 2019)},
  volume = {19},
  year = {2019},
  _bib2doi_finished = {true},
}

@article{heule2019sat,
author = {Marijn J. H. Heule and Matti Järvisalo and Martin Suda},
title ={SAT Competition 2018},
journal = {Journal on Satisfiability, Boolean Modelling and Computation},
volume = {11},
number = {1},
pages = {133-154},
year = {2019},
doi = {10.3233/SAT190120},
URL = { https://journals.sagepub.com/doi/abs/10.3233/SAT190120},
eprint = { https://journals.sagepub.com/doi/pdf/10.3233/SAT190120}
}

@article{hnich2006constraint,
  title={Constraint models for the covering test problem},
  author={Hnich, Brahim and Prestwich, Steven D and Selensky, Evgeny and Smith, Barbara M},
  journal={Constraints},
  volume={11},
  number={2},
  pages={199--219},
  year={2006},
  publisher={Springer}
}

@inproceedings{
hou2026llmbranch,
title={{LLM}4Branch: Large Language Model for Discovering Efficient Branching Policies of Integer Programs},
author={Zhinan Hou and Xingchen Li and Yankai Zhang and Tianxun Li and Keyou You},
booktitle={Forty-third International Conference on Machine Learning},
year={2026},
url={https://openreview.net/forum?id=YIzUTHEvG7}
}

@inproceedings{ijcai2023p211,
  title     = {Learning When to Use Automatic Tabulation in Constraint Model Reformulation},
  author    = {Cena, Carlo and Akg{\"u}n, {\"O}zg{\"u}r and Kiziltan, Zeynep and Miguel, Ian and Nightingale, Peter and Ulrich-Oltean, Felix},
  booktitle = {Proceedings of the Thirty-Second International Joint Conference on
               Artificial Intelligence},
  pages     = {1902--1910},
  year      = {2023},
  month     = {8},
  note      = {Main Track},
  doi       = {10.24963/ijcai.2023/211},
  url       = {https://doi.org/10.24963/ijcai.2023/211},
}

@inproceedings{ishay_leveraging_2023,
  title={Leveraging Large Language Models to Generate Answer Set Programs},
  author={Ishay, Adam and Yang, Zhun and Lee, Joohyung},
  booktitle={Proceedings of the International Conference on Principles of Knowledge Representation and Reasoning},
  volume={19},
  pages={374--383},
  year={2023}
}

@inproceedings{liu2019solving,
  title={Solving the Social Golfers Problems by Constraint Programming in Sequential and Parallel.},
  author={Liu, Ke and L{\"o}ffler, Sven and Hofstedt, Petra},
  booktitle={ICAART (2)},
  pages={29--39},
  year={2019}
}

@article{liu2024llm4ad,
    title = {LLM4AD: A Platform for Algorithm Design with Large Language Model},
    author = {Fei Liu and Rui Zhang and Zhuoliang Xie and Rui Sun and Kai Li and Xi Lin and Zhenkun Wang and Zhichao Lu and Qingfu Zhang},
    year = {2024},
    eprint = {2412.17287},
    archivePrefix = {arXiv},
    primaryClass = {cs.AI},
    url = {https://arxiv.org/abs/2412.17287},
}

@article{liu2026systematic,
  title={A systematic survey on large language models for algorithm design},
  author={Liu, Fei and Yao, Yiming and Guo, Ping and Yang, Zhiyuan and Lin, Xi and Zhao, Zhe and Tong, Xialiang and Mao, Kun and Lu, Zhichao and Wang, Zhenkun and others},
  journal={ACM Computing Surveys},
  volume={58},
  number={8},
  pages={1--32},
  year={2026},
  publisher={ACM New York, NY}
}

@inproceedings{liuevolution,
  title={Evolution of Heuristics: Towards Efficient Automatic Algorithm Design Using Large Language Model},
  author={Liu, Fei and Xialiang, Tong and Yuan, Mingxuan and Lin, Xi and Luo, Fu and Wang, Zhenkun and Lu, Zhichao and Zhang, Qingfu},
  booktitle={Forty-first International Conference on Machine Learning},
  year={2024}
}

@article{meyerson2024language,
  title={Language model crossover: Variation through few-shot prompting},
  author={Meyerson, Elliot and Nelson, Mark J and Bradley, Herbie and Gaier, Adam and Moradi, Arash and Hoover, Amy K and Lehman, Joel},
  journal={ACM Transactions on Evolutionary Learning},
  volume={4},
  number={4},
  pages={1--40},
  year={2024},
  publisher={ACM New York, NY}
}

@inproceedings{michailidis2024constraint,
  title={Constraint modelling with LLMs using in-context learning},
  author={Michailidis, Kostis and Tsouros, Dimos and Guns, Tias},
  booktitle={30th International conference on principles and practice of constraint programming},
  year={2024}
}

@inproceedings{michailidis2025cp,
  title={CP-Bench: Evaluating Large Language Models for Constraint Modelling},
  author={Michailidis, Kostis and Tsouros, Dimos and Guns, Tias},
  booktitle={28th European Conference on Artificial Intelligence (ECAI)},
  year={2025}
}

@inproceedings{minizinc,
  title={MiniZinc: Towards a standard CP modelling language},
  author={Nethercote, Nicholas and Stuckey, Peter J and Becket, Ralph and Brand, Sebastian and Duck, Gregory J and Tack, Guido},
  booktitle={International conference on principles and practice of constraint programming},
  pages={529--543},
  year={2007},
  organization={Springer}
}

@inproceedings{mmr,
  author = {Jaime G. Carbonell and Jade Goldstein},
  title = {The Use of MMR, Diversity-Based Reranking for Reordering Documents and Producing Summaries},
  year = {1998},
  isbn = {1581130155},
  publisher = {{ACM}},
  address = {New York, NY, USA},
  url = {https://doi.org/10.1145/290941.291025},
  doi = {10.1145/290941.291025},
  booktitle = {{SIGIR} '98: Proceedings of the 21st Annual International {ACM} {SIGIR} Conference on Research and Development in Information Retrieval, August 24-28 1998, Melbourne, Australia},
  pages = {335--336},
  numpages = {2},
  location = {Melbourne, Australia},
  series = {SIGIR '98},
  timestamp = {Wed, 14 Nov 2018 10:58:11 +0100},
  biburl = {https://dblp.org/rec/conf/sigir/CarbonellG98.bib},
  bibsource = {dblp computer science bibliography, https://dblp.org},
  editor = {W. Bruce Croft and Alistair Moffat and C. J. van Rijsbergen and Ross Wilkinson and Justin Zobel},
  _bib2doi_selected = {dblp:/rec/conf/sigir/CarbonellG98.bib},
  _bib2doi_confirmed = {true},
  _bib2doi_finished = {true},
}

@inproceedings{moffitt2006optimal,
  title={Optimal Rectangle Packing: A Meta-CSP Approach.},
  author={Moffitt, Michael D and Pollack, Martha E},
  booktitle={ICAPS},
  pages={93--102},
  year={2006}
}

@article{nightingale2017automatically,
  title={Automatically improving constraint models in Savile Row},
  author={Nightingale, Peter and Akg{\"u}n, {\"O}zg{\"u}r and Gent, Ian P and Jefferson, Christopher and Miguel, Ian and Spracklen, Patrick},
  journal={Artificial Intelligence},
  volume={251},
  pages={35--61},
  year={2017},
  publisher={Elsevier}
}

@misc{ortools,
  title = {OR-Tools},
  version = { v9.10 },
  author = {Laurent Perron and Vincent Furnon},
  organization = {Google},
  url = {https://developers.google.com/optimization/},
note="Accessed: 2025-5-5",
year={2024}
}

@inproceedings{regin1996generalized,
  title={Generalized arc consistency for global cardinality constraint},
  author={R{\'e}gin, Jean-Charles},
  booktitle={Proceedings of the thirteenth national conference on Artificial intelligence-Volume 1},
  pages={209--215},
  year={1996}
}

@inproceedings{schaus2008global,
  title={A Global Constraint for Bin-Packing with Precedences: Application to the Assembly Line Balancing Problem.},
  author={Schaus, Pierre and Deville, Yves and others},
  booktitle={AAAI},
  pages={369--374},
  year={2008}
}

@inproceedings{shi-etal-2025-constraintllm,
    title = "{C}onstraint{LLM}: A Neuro-Symbolic Framework for Industrial-Level Constraint Programming",
    author = "Shi, Weichun  and
      Liu, Minghao  and
      Zhang, Wanting  and
      Shi, Langchen  and
      Jia, Fuqi  and
      Ma, Feifei  and
      Zhang, Jian",
    booktitle = "Proceedings of the 2025 Conference on Empirical Methods in Natural Language Processing",
    month = nov,
    year = "2025",
    address = "Suzhou, China",
    url = "https://aclanthology.org/2025.emnlp-main.809/",
    doi = "10.18653/v1/2025.emnlp-main.809",
    pages = "16010--16030",
}

@inproceedings{simonis1999building,
  title = {Building Industrial Applications with Constraint Programming},
  author = {Helmut Simonis},
  booktitle = {Constraints in Computational Logics: Theory and Applications, International Summer School, CCL'99 Gif-sur-Yvette, France, September 5-8, 1999, Revised Lectures},
  pages = {271--309},
  year = {1999},
  publisher = {Springer},
  bibsource = {dblp computer science bibliography, https://dblp.org},
  doi = {10.1007/3-540-45406-3_6},
  volume = {2002},
  series = {Lecture Notes in Computer Science}
}

@inproceedings{song2025llmcp,
  title={Do LLMs Understand Constraint Programming? Zero-Shot Constraint Programming Model Generation Using LLMs},
  author={Song, Yuliang and Cohen, Eldan},
  booktitle={Proceedings of the 19th Learning and Intelligent Optimization Conference (LION-25)},
  pages={In press},
  year={2025},
  url={https://openreview.net/forum?id=6zlpzSKzqj}
}

@article{spracklen2023automated,
  title={Automated streamliner portfolios for constraint satisfaction problems},
  author={Spracklen, Patrick and Dang, Nguyen and Akg{\"u}n, {\"O}zg{\"u}r and Miguel, Ian},
  journal={Artificial Intelligence},
  volume={319},
  pages={103915},
  year={2023},
  publisher={Elsevier}
}

@article{sun2026autosat,
  author  = {Yiwen Sun and Furong Ye and Xianyin Zhang and Shiyu Huang and
             Bingzhen Zhang and Ke Wei and Shaowei Cai},
  title   = {{AutoSAT}: Automatically Optimize {SAT} Solvers via Large
             Language Models},
  journal = {Journal of Artificial Intelligence Research},
  volume  = {86},
  pages   = {29:1--29:50},
  month   = jul,
  year    = {2026},
  doi     = {10.1613/jair.1.20499}
}

@article{van2024llamea,
  title={Llamea: A large language model evolutionary algorithm for automatically generating metaheuristics},
  author={Van Stein, Niki and B{\"a}ck, Thomas},
  journal={IEEE Transactions on Evolutionary Computation},
  volume={29},
  number={2},
  pages={331--345},
  year={2024},
  publisher={IEEE}
}

@article{vaswani_attention_2023,
  title={Attention is all you need},
  author={Vaswani, Ashish and Shazeer, Noam and Parmar, Niki and Uszkoreit, Jakob and Jones, Llion and Gomez, Aidan N and Kaiser, {\L}ukasz and Polosukhin, Illia},
  journal={Advances in neural information processing systems},
  volume={30},
  year={2017}
}

@article{voboril2025generating,
  title={Generating streamlining constraints with large language models},
  author={Voboril, Florentina and Ramaswamy, Vaidyanathan Peruvemba and Szeider, Stefan},
  journal={Journal of Artificial Intelligence Research},
  volume={84},
  year={2025}
}

@inproceedings{
wang2023selfconsistency,
title={Self-Consistency Improves Chain of Thought Reasoning in Language Models},
author={Xuezhi Wang and Jason Wei and Dale Schuurmans and Quoc V Le and Ed H. Chi and Sharan Narang and Aakanksha Chowdhery and Denny Zhou},
booktitle={The Eleventh International Conference on Learning Representations },
year={2023},
url={https://openreview.net/forum?id=1PL1NIMMrw}
}

@inproceedings{xiao2025survey,
  title={A survey of optimization modeling meets LLMs: progress and future directions},
  author={Xiao, Ziyang and Xie, Jingrong and Xu, Lilin and Guan, Shisi and Zhu, Jingyan and Han, Xiongwei and Fu, Xiaojin and Yu, WingYin and Wu, Han and Shi, Wei and others},
  booktitle={Proceedings of the Thirty-Fourth International Joint Conference on Artificial Intelligence},
  pages={10742--10750},
  year={2025}
}

@article{xu2026deepseek,
  title={Deepseek-v4: Towards highly efficient million-token context intelligence},
  author={Xu, Anyi and Lin, Bangcai and Xue, Bing and Wang, Bingxuan and Xu, Bingzheng and Wu, Bochao and Zhang, Bowei and Lin, Chaofan and Dong, Chen and Ling, Chenchen and others},
  journal={arXiv preprint arXiv:2606.19348},
  year={2026}
}

@article{ye_satlm_2023,
  title={Satlm: Satisfiability-aided language models using declarative prompting},
  author={Ye, Xi and Chen, Qiaochu and Dillig, Isil and Durrett, Greg},
  journal={Advances in Neural Information Processing Systems},
  volume={36},
  pages={45548--45580},
  year={2023}
}

@inproceedings{ye2024reevo,
    title={ReEvo: Large Language Models as Hyper-Heuristics with Reflective Evolution}, 
    author={Haoran Ye and Jiarui Wang and Zhiguang Cao and Federico Berto and Chuanbo Hua and Haeyeon Kim and Jinkyoo Park and Guojie Song},
    booktitle={Advances in Neural Information Processing Systems},
    year={2024}
}

@inproceedings{
zhang2026rethinking,
title={Rethinking Code Similarity for Automated Algorithm Design with {LLM}s},
author={Rui Zhang and Zhichao Lu},
booktitle={The Fourteenth International Conference on Learning Representations},
year={2026},
url={https://openreview.net/forum?id=HIUqeO9OOr}
}

\FloatBarrier
\appendix
\section{Prompt, Generation, and Evaluation Details}
\label{app:prompt-details}
\label{app:evaluation-details}

All ten reformulation configurations use the following system instruction.
The target solver, fixed training-cap rule, and complete baseline model are
appended at runtime. The first user message lists the baseline training
runtimes; retained assistant answers and measured feedback are inserted as
alternating assistant and user messages; and the final user message contains
the task instruction.

\begin{lstlisting}[
  style=supplementlisting,
  caption={Shared reformulation system prompt.},
  label={lst:reformulation-system-prompt}
]
## Role
You are an expert in constraint programming.
Your objective is to reformulate a given CPMpy baseline model to maximize solving efficiency, while strictly preserving semantic correctness.
A successful reformulation must be: (1) semantically equivalent to the baseline, and (2) faster for the target solver than the baseline or any previously discovered valid candidate.

## Output Format
Your visible answer must contain exactly two top-level sections, in this order:
```
## Rationale
Provide a technical analysis of the reformulation strategy. Explain why semantic equivalence is maintained and provide a theoretical justification for the expected performance improvement.
## Final Model
One fenced `python` code block with the complete model.
```

## Rules
1. Keep the function signature build_model(data) -> (model, extractor) and the same input contract (data keys and meaning) as the baseline.
2. You may change decision variables or viewpoints freely, but extractor() must return the exact same JSON structure as the baseline (same keys and value types); how you map your variables to it is up to you.
3. Preserve semantics exactly: the model must not accept invalid solutions and must stay satisfiable for any instance possible (even out of this training set).
4. Dependencies: `import cpmpy as cp`, numpy, and the Python standard library only. Prefer `cp.sum`, `cp.max`, `cp.min`, `cp.any`, `cp.all` over Python built-ins.
5. Integer modeling only (floats/fractional constants are strictly not allowed); no printing, no file I/O, no extra scripts, no solving.
6. Treat prior proposals as explored candidates; do not repeat a previous model with only superficial changes.

## CPMpy Documentation

CPMpy is a Constraint Programming and Modeling library in Python, based on numpy.

### Model
- `cp.Model(*constraints)`: create a model.
- `model += constraint`: add a constraint.
- `model.minimize(expr)` / `model.maximize(expr)`: set optimization objective.

### Variables
- `cp.boolvar(shape=None)`: Boolean decision variable(s).
- `cp.intvar(lb, ub, shape=None)`: Integer decision variable(s).
- `cp.cpm_array(arr)`: CPMpy-aware array helper.
- Values: `x.value()`, arrays via `x.value().tolist()`.

### Core expressions / operators
- Comparisons: `==`, `!=`, `<`, `<=`, `>`, `>=`.
- Arithmetic: `+`, `-`, `*`, `//`, `%`, `**`, unary `-`.
- CAUTION (broadcasting): unlike NumPy, `*` between arrays of different rank does not broadcast reliably (e.g. a rank-2 array times a rank-1 vector silently mis-aligns elements). For such products use `np.multiply(a, b)` or explicit indexing.
- Logic: `&`, `|`, `~`, `^`, implication `a.implies(b)`.
- Aggregations: `cp.sum`, `cp.abs`, `cp.max`, `cp.min`, `cp.all`, `cp.any`.
- Indexing with decision vars: use `x[idx]`; for Python lists, wrap with `cp.cpm_array(...)` first.

### Global constraints
- `cp.AllDifferent(*args)`: All arguments have different values.
- `cp.AllDifferentExcept0(*args)`: All nonzero arguments have different values.
- `cp.AllDifferentExceptN(args, n)`: All arguments except those equal to a value in 'n' have a distinct value.
- `cp.AllEqual(*args)`: All arguments have the same value.
- `cp.AllEqualExceptN(args, n)`: All arguments except those equal to a value in 'n' have the same value.
- `cp.Circuit(*args)`: Variables form a circuit (e.g., for routing), where x[i] = j means that j is the successor of i.
- `cp.Inverse(array1, array2)`: Inverse (aka channeling / assignment) constraint. The value of array1[i] is the index of the value in array2.
- `cp.Table(array, table)`: Variables in array must match a row in table.
- `cp.ShortTable(array, table)`: Extension of the `Table` constraint where the `table` matrix may contain wildcards, meaning there are no restrictions for the corresponding variable in that tuple.
- `cp.NegativeTable(array, table)`: The values of the variables in 'array' do not correspond to any row in 'table'.
- `cp.IfThenElse(cond, if_true, if_false)`: If cond is true, then if_true else if_false. All arguments must be boolean expressions.
- `cp.InDomain(expr, domain)`: Defines non-interval domains for an expression.
- `cp.Xor(arg_list)`: Exclusive-or constraint for the arguments.
- `cp.Cumulative(start, duration, end, demand, capacity)`: Ensures no task overlaps and respects resource capacity.
- `cp.Precedence(vars, p)`: Constraint enforcing some values have precedence over others. If vars[i] = p[j+1], then there exists a vars[i'] = p[j] with i' < i
- `cp.NoOverlap(start, dur, end)`: Ensures that the intervals defined by start, dur, and end do not overlap.
- `cp.GlobalCardinalityCount(vars, vals, occ)`: Specifies the number of occurrences of values in a variable list.
- `cp.Increasing(array)`: The elements of array are in non-decreasing order.
- `cp.IncreasingStrict(array)`: Same as Increasing, but strictly increasing.
- `cp.Decreasing(array)`: The elements of array are in non-increasing order.
- `cp.DecreasingStrict(array)`: Same as Decreasing, but strictly decreasing.
- `cp.LexLess(l1, l2)`: List l1 is lexicographically less than list l2.
- `cp.LexLessEq(l1, l2)`: List l1 is lexicographically less than or equal to list l2.
- `cp.LexChainLess(X)`: All rows of matrix X are lexicographically ordered.
- `cp.LexChainLessEq(X)`: Same as LexChainLess, but with less than or equal.
- `cp.DirectConstraint(name, arguments, novar=None)`: Directly call a solver-specific global constraint by name. Arguments should be a tuple of arguments to pass to the solver function with name 'name'. novar is a list of indices (offset 0) of arguments in `arguments` that contain no variables, that can be passed 'as is' without scanning for variables.

### Global functions
- `cp.Minimum(arg_list)`: Computes the minimum value of the arguments.
- `cp.Maximum(arg_list)`: Computes the maximum value of the arguments.
- `cp.Abs(expr)`: Computes the absolute value of an expression.
- `cp.Element(arr, idx)`: Enforces arr[idx] to match a specific result. It is generally better to use `arr[idx]` directly.
- `cp.Count(arr, val)`: Counts occurrences of val in arr.
- `cp.Among(arr, vals)`: Counts variables taking values in vals.
- `cp.NValue(arr)`: Counts distinct values in arr.
- `cp.NValueExcept(arr, n)`: Counts distinct values in arr excluding n.
\end{lstlisting}

The base instruction used by the non-operator strategies and EoH
initialisation is exact. Before any retained attempts, the first user message
uses this runtime table:
\begin{lstlisting}[
  style=supplementlisting,
  caption={Baseline-runtime user-message template.},
  label={lst:baseline-runtime-prompt}
]
## Baseline Training Runtimes

| Training bin | Mean baseline time (s) |
|---:|---:|
| <bin> | <seconds> |
\end{lstlisting}

\begin{lstlisting}[
  style=supplementlisting,
  caption={Base reformulation task instruction.},
  label={lst:base-task-prompt}
]
## Task

Propose one materially new faster equivalent model.
\end{lstlisting}

Operators $e_1$ and $e_2$ receive two parents; $m_1$ and $m_2$ receive one.
Each operator text is placed below the same \texttt{\#\# Task} heading.
Bracketed labels below are descriptive and are not part of the prompt.
\begin{lstlisting}[
  style=supplementlisting,
  caption={EoH operator instructions. Bracketed labels are not sent to the model.},
  label={lst:eoh-operator-prompts}
]
[e1: distinct approach]
Create a new semantically equivalent reformulation that takes a fundamentally different modelling approach from each parent model above. Avoid superficial syntax changes. The reformulation should be expected to improve solving efficiency over the original baseline.

[e2: compatible combination]
Create a new semantically equivalent reformulation using common or compatible ideas from the parent models above. You may either combine compatible parent components, extend the shared ideas, or instantiate them in a substantially different way. Do not blindly concatenate parent models. The reformulation should be expected to improve solving efficiency over the original baseline.

[m1: structural modification]
Create a new semantically equivalent reformulation by making a local structural change to the model given above while preserving its main reformulation idea. The reformulation should be expected to improve solving efficiency over the given parent for the target solver.

[m2: component modification]
Create a new semantically equivalent reformulation by modifying one specific solver-facing component of the model given above, while keeping the main model structure unchanged. The reformulation should be expected to improve solving efficiency over the given parent for the target solver.
\end{lstlisting}

A retained attempt is followed by one of the feedback forms below. Values in
angle brackets are filled from its training evaluation. EoH omits measured
feedback for valid parents, whereas EoH+feedback includes it.
\begin{lstlisting}[
  style=supplementlisting,
  caption={Measured-feedback forms. Optional lines are shown in brackets.},
  label={lst:feedback-prompts}
]
## Measured Performance

Result: valid - solved <n>/<N> training instances.
Total time: <candidate>s vs baseline <baseline>s (speedup <ratio>x).
[Per-instance: faster on <n>/<N>, slower on <n>/<N> (...).]

Result: capped - evaluation stopped at the iteration cap (<cap>s) after solving <n>/<N> instances; true total runtime unknown, so it did not beat the incumbent that set the cap.
[Solved subset: <candidate>s vs baseline <baseline>s on the <n> solved instances (<relation>).]

Result: stopped early - <instance(s)> hit the per-instance timeout with <n>/<N> instances solved; the total runtime is not comparable to the baseline.
[Solved subset: <candidate>s vs baseline <baseline>s on the <n> solved instances (<relation>).]

Result: invalid (<class>) - <description>. Solved <n>/<N> instances before stopping; candidate rejected.
[Error: <detail>]

Result: not evaluated - no runnable model could be parsed from the answer (<class>).
[Error: <detail>]
\end{lstlisting}

When EoH has too few valid models, recent rejected attempts fill the missing
context slots but cannot be selected as parents. Incomplete offspring batches
do not trigger survival. Parent sampling is seeded by the iteration, and
runtime ties are resolved by iteration and model identifier.

One recovery call is made with the original transcript and partial response:
\begin{lstlisting}[
  style=supplementlisting,
  caption={Truncated-answer recovery instruction.},
  label={lst:recovery-prompt}
]
## Role and Goal

You are completing a CPMpy reformulation answer that was truncated before a complete visible rationale and final model were provided.

## Critical Instructions

1) Use the baseline model, earlier conversation, measured feedback, and partial reasoning as context.
2) Output exactly two top-level sections: `## Rationale` and `## Final Model`.
3) Under `## Rationale`, provide a concise visible rationale based on the partial answer.
4) Under `## Final Model`, output exactly one fenced `python` code block containing a complete standalone implementation.
5) Keep the same function signature: build_model(data) -> (model, extractor).
6) Preserve the same input contract and exact extractor output schema as the baseline.
7) Do not include additional sections, markdown tables, or extra code blocks.
\end{lstlisting}

\paragraph{Execution details not given in the main paper.}
Instance parameter values are not included in prompts. Solver seed 42 is fixed,
and the returned-answer check has a 60-second limit. The baseline is remeasured
before each search. An instance is excluded from that run if the live baseline
times out, cannot be evaluated, or cannot complete the returned-answer check.
Provider failures are retried up to three consecutive times and never enter
history. Parse and evaluation failures do enter history, while only completed
solution-valid candidates can become training incumbents. Exact request
messages, normalized responses and reasoning, token use, generation settings,
parsed candidates, evaluation summaries, and separate provider-failure records
are retained without duplicating prompts.

\section{Analysis and Frozen Split}
\label{app:analysis-details}
\label{app:instance-split}

\begin{figure*}[p]
\centering
\includegraphics[width=0.85\textwidth]{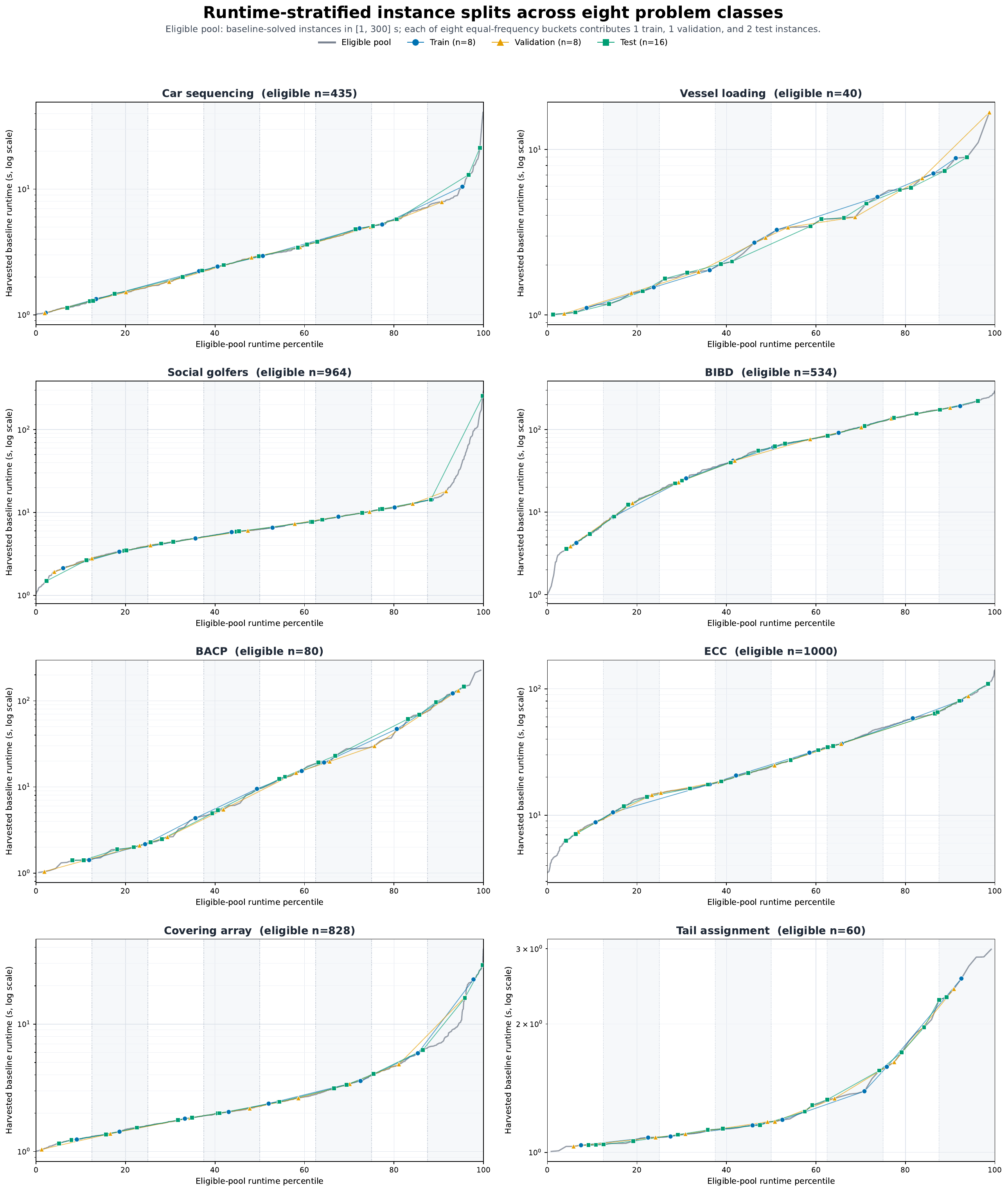}
\caption{Baseline-runtime distributions and selected instances under the
runtime-stratified protocol. Each panel shows the eligible pool and selected
training, validation, and test instances.}
\label{fig:supp-split-runtimes}
\end{figure*}

\paragraph{Failure-aware scoring.}
A selected model that completes the retained test set and passes the
returned-answer check keeps its measured speedup. Each per-instance timeout
contributes twice the 300-second limit.
Figure~\ref{fig:supp-split-runtimes} shows the eligible baseline-runtime
distribution and the selected split for each problem.

\section{Additional Results}
\label{app:additional-results}

\begin{figure*}[tb]
\centering
\includegraphics[width=0.89\textwidth]{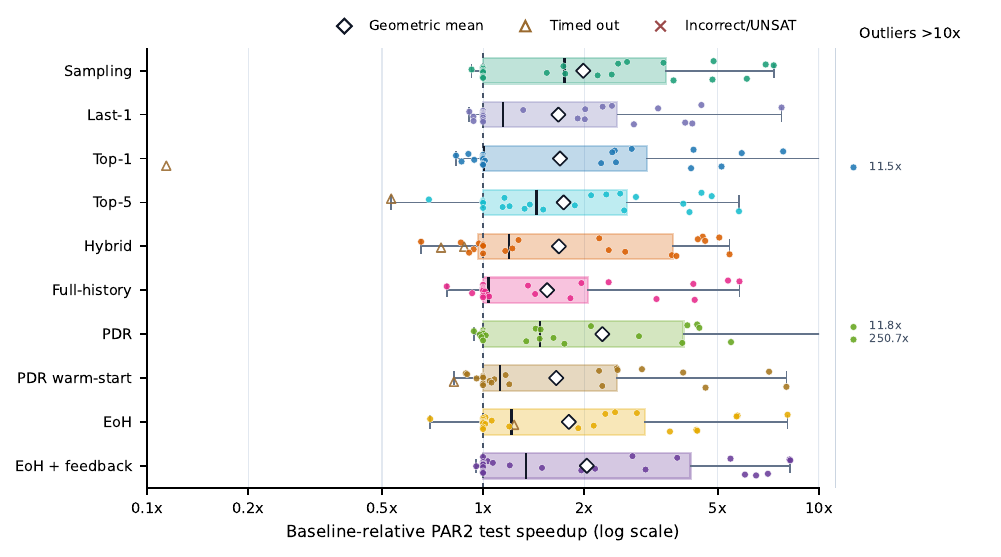}
\caption{Distribution of BR-PAR2 test outcomes. Each marker is one search;
diamonds mark geometric means and the right gutter lists outliers above
$10\times$.}
\label{fig:supp-distribution}
\end{figure*}

Table~\ref{tab:supp-problem-gm} gives the full per-problem matrix. The outcome
distribution in Figure~\ref{fig:supp-distribution} includes models that
validate faster but run slower on test.

\begin{table*}[tb]
\centering
\resizebox{\linewidth}{!}{%
\begin{tabular}{lrrrrrrrr}
\toprule
Strategy & Car sequencing & Vessel loading & Social golfers & BIBD & BACP & ECC & Covering array & Tail assignment \\
\midrule
Sampling & 3.55 & 5.04 & \textbf{1.20} & 0.97 & 1.16 & 2.10 & 1.00 & 4.77 \\
Last-1 & 2.90 & 3.72 & 1.06 & 1.00 & 0.98 & 1.57 & 0.98 & 3.61 \\
Top-1 & 3.07 & 4.81 & 0.91 & 1.00 & 0.98 & 1.69 & 0.48 & 6.27 \\
Top-5 & 2.98 & 3.06 & 0.75 & 1.00 & \textbf{1.76} & 1.37 & 1.26 & 3.97 \\
Hybrid & 3.69 & 2.94 & 0.82 & 1.00 & 0.93 & 1.37 & 1.34 & 4.20 \\
Full-history & \textbf{4.19} & 2.02 & 1.02 & 1.00 & 1.17 & 1.11 & 1.01 & 2.94 \\
PDR & 3.33 & 4.21 & 1.12 & 1.00 & 1.43 & 1.33 & \textbf{6.96} & 3.34 \\
PDR warm-start & 2.65 & 3.96 & 0.98 & 1.00 & 1.20 & 0.97 & 1.06 & 4.36 \\
EoH & 2.54 & 6.40 & 0.89 & 1.02 & 1.00 & 1.47 & 1.24 & 4.07 \\
EoH + feedback & 2.19 & \textbf{7.15} & 1.03 & \textbf{1.06} & 1.13 & \textbf{2.34} & 1.00 & \textbf{6.51} \\
\bottomrule
\end{tabular}
}

\caption{BR-PAR2 test GM per problem and strategy over three repetitions. Bold
marks the best strategy per problem.}
\label{tab:supp-problem-gm}
\end{table*}

Table~\ref{tab:supp-matched-comparisons} gives the paired comparisons used for
the ablation results reported in the main paper.
Table~\ref{tab:supp-search-diagnostics} summarises proposal validity, token use,
and cost by strategy.

\begin{table}[tb]
\centering
\scriptsize
\resizebox{\linewidth}{!}{%
\begin{tabular}{lrrr}
\toprule
Contrast & \shortstack{GM ratio\\{}[95\% CI]} & \shortstack{Better /\\tie / worse} & Median \\
\midrule
Top-1 vs Last-1 & 1.01$\times$ [0.79, 1.25] & 11 / 6 / 7 & 1.00$\times$ \\
Hybrid vs Top-1 & 0.99$\times$ [0.77, 1.39] & 8 / 3 / 13 & 0.93$\times$ \\
Hybrid vs Last-1 & 1.00$\times$ [0.88, 1.15] & 9 / 4 / 11 & 1.00$\times$ \\
PDR vs Top-5 & 1.30$\times$ [0.95, 2.05] & 10 / 3 / 11 & 1.00$\times$ \\
EoH + feedback vs EoH & 1.13$\times$ [0.97, 1.34] & 14 / 5 / 5 & 1.16$\times$ \\
PDR vs PDR warm-start & 1.37$\times$ [1.00, 2.23] & 12 / 3 / 9 & 1.01$\times$ \\
\bottomrule
\end{tabular}
}

\caption{Matched GM speedup comparisons over problem--repetition cells.}
\label{tab:supp-matched-comparisons}
\end{table}

\begin{table}[tb]
\centering
\scriptsize
\resizebox{\linewidth}{!}{%
\begin{tabular}{lrrrrr}
\toprule
Policy & \shortstack{Strict-valid\\proposals} & \shortstack{Cached\\input} & \shortstack{Uncached\\input} & \shortstack{Output} & \shortstack{Cost\\per run} \\
\midrule
Sampling & 31.5\% & 0.245M & 0.010M & 0.598M & \$0.17 \\
Last-1 & 34.9\% & 0.245M & 0.107M & 0.412M & \$0.13 \\
Top-1 & 33.2\% & 0.333M & 0.013M & 0.456M & \$0.13 \\
Top-5 & 40.2\% & 0.580M & 0.069M & 0.419M & \$0.13 \\
Hybrid & 40.4\% & 0.326M & 0.109M & 0.386M & \$0.12 \\
Full-history & 34.9\% & 6.575M & 0.160M & 0.292M & \$0.12 \\
PDR & 42.0\% & 0.579M & 0.062M & 0.406M & \$0.12 \\
PDR warm-start & 45.5\% & 0.603M & 0.077M & 0.416M & \$0.13 \\
EoH & 42.8\% & 0.336M & 0.068M & 0.491M & \$0.15 \\
EoH + feedback & 44.0\% & 0.342M & 0.074M & 0.496M & \$0.15 \\
\bottomrule
\end{tabular}
}

\caption{Search diagnostics by strategy.}
\label{tab:supp-search-diagnostics}
\end{table}

\section{Highest Speedup Example: Covering Arrays}
\label{app:example-models}

This is the largest observed gain (PDR, $250.7\times$ on test).
The baseline states coverage directly: for every $t$-subset of rows and every
value tuple, some column must match. The selected model encodes each column on
a row subset as one base-$g$ integer and forces the first $g^t$ columns to
enumerate all tuples in increasing order. This restricts the returned covering
arrays. Every training, validation, and test answer returned by this model was
accepted by the baseline.
Listings~\ref{lst:covering-baseline}
and~\ref{lst:covering-pdr} show the initial baseline and the complete generated
model.

\begin{lstlisting}[
  style=supplementlisting,
  language=Python,
  caption={Initial covering-array baseline.},
  label={lst:covering-baseline}
]
import cpmpy as cp
import itertools


def build_model(data):
    t = data["t"]
    k = data["k"]
    g = data["g"]
    b = data["b"]

    matrix = cp.intvar(0, g - 1, shape=(k, b), name="matrix")
    model = cp.Model()

    row_combinations = list(itertools.combinations(range(k), t))
    value_tuples = list(itertools.product(range(g), repeat=t))

    for rows in row_combinations:
        for vals in value_tuples:
            col_matches = []
            for c in range(b):
                col_matches.append(
                    cp.all(
                        [
                            matrix[rows[i], c] == vals[i]
                            for i in range(t)
                        ]
                    )
                )
            model += cp.any(col_matches)

    def extractor():
        return {"matrix": matrix.value().tolist()}

    return model, extractor
\end{lstlisting}

\begin{lstlisting}[
  style=supplementlisting,
  language=Python,
  caption={Covering arrays: PDR repetition 1, iteration 23.},
  label={lst:covering-pdr}
]
import cpmpy as cp
import itertools


def build_model(data):
    """
    Build a covering array model using integer encoding and strict ordering
    on the first g^t columns to enforce covering and break column symmetry.
    """
    t = data["t"]
    k = data["k"]
    g = data["g"]
    b = data["b"]

    matrix = cp.intvar(0, g - 1, shape=(k, b), name="matrix")
    model = cp.Model()

    # Precompute powers of g
    g_pows = [g ** i for i in range(t)]
    g_pow_t = g ** t

    # All combinations of t rows
    row_combos = list(itertools.combinations(range(k), t))

    for rows in row_combos:
        # Encoded tuple per column
        comb = cp.intvar(0, g_pow_t - 1, shape=b, name=f"comb_{rows}")

        for c in range(b):
            model += (comb[c] == cp.sum(matrix[rows[i], c] * g_pows[i] for i in range(t)))

        # For the first g^t columns, enforce increasing order.
        # Because there are exactly g^t values from a domain of size g^t,
        # this forces the sequence to be [0,1,...,g^t-1], thereby covering
        # every tuple exactly once.
        if b >= g_pow_t:
            for c in range(g_pow_t - 1):
                model += (comb[c] < comb[c + 1])

    def extractor():
        return {"matrix": matrix.value().tolist()}

    return model, extractor
\end{lstlisting}

\section{Per-Run Search Trajectories}
\label{app:trajectories}

\begin{figure*}[p]
\centering
\includegraphics[width=\textwidth]{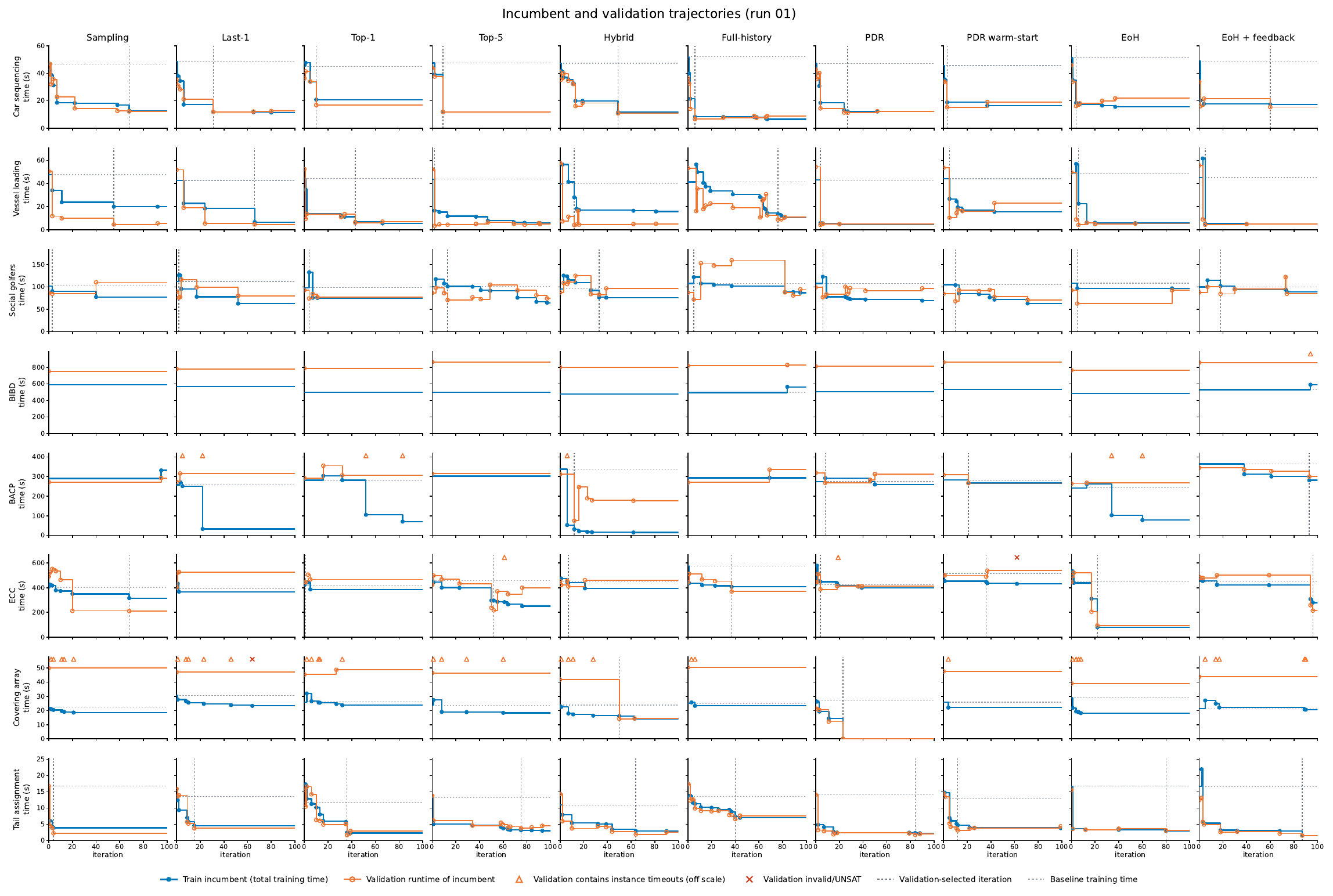}
\caption{Incumbent and validation trajectories, repetition 1.}
\label{fig:supp-grid-1}
\end{figure*}

\begin{figure*}[p]
\centering
\includegraphics[width=\textwidth]{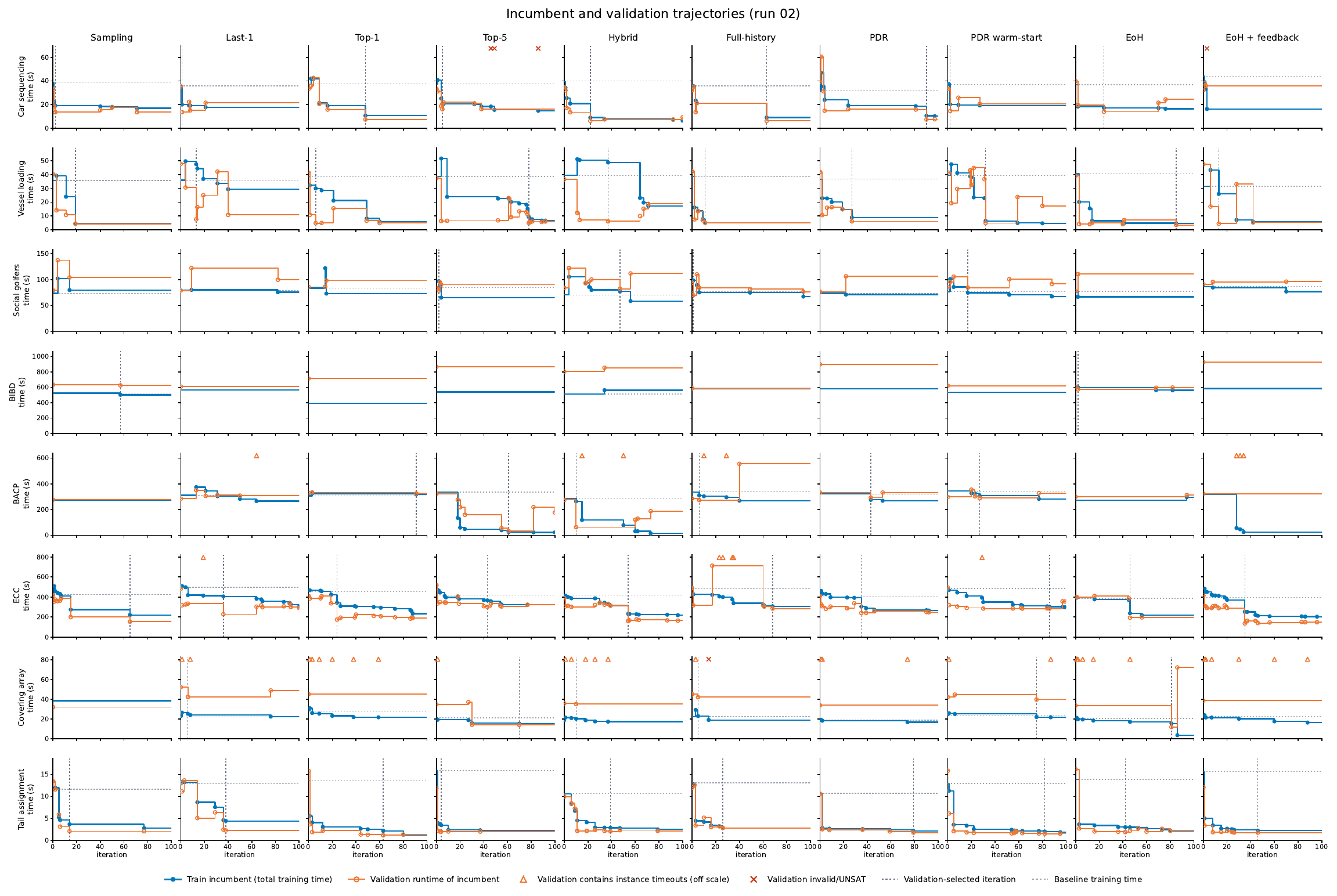}
\caption{Incumbent and validation trajectories, repetition 2.}
\label{fig:supp-grid-2}
\end{figure*}

\begin{figure*}[p]
\centering
\includegraphics[width=\textwidth]{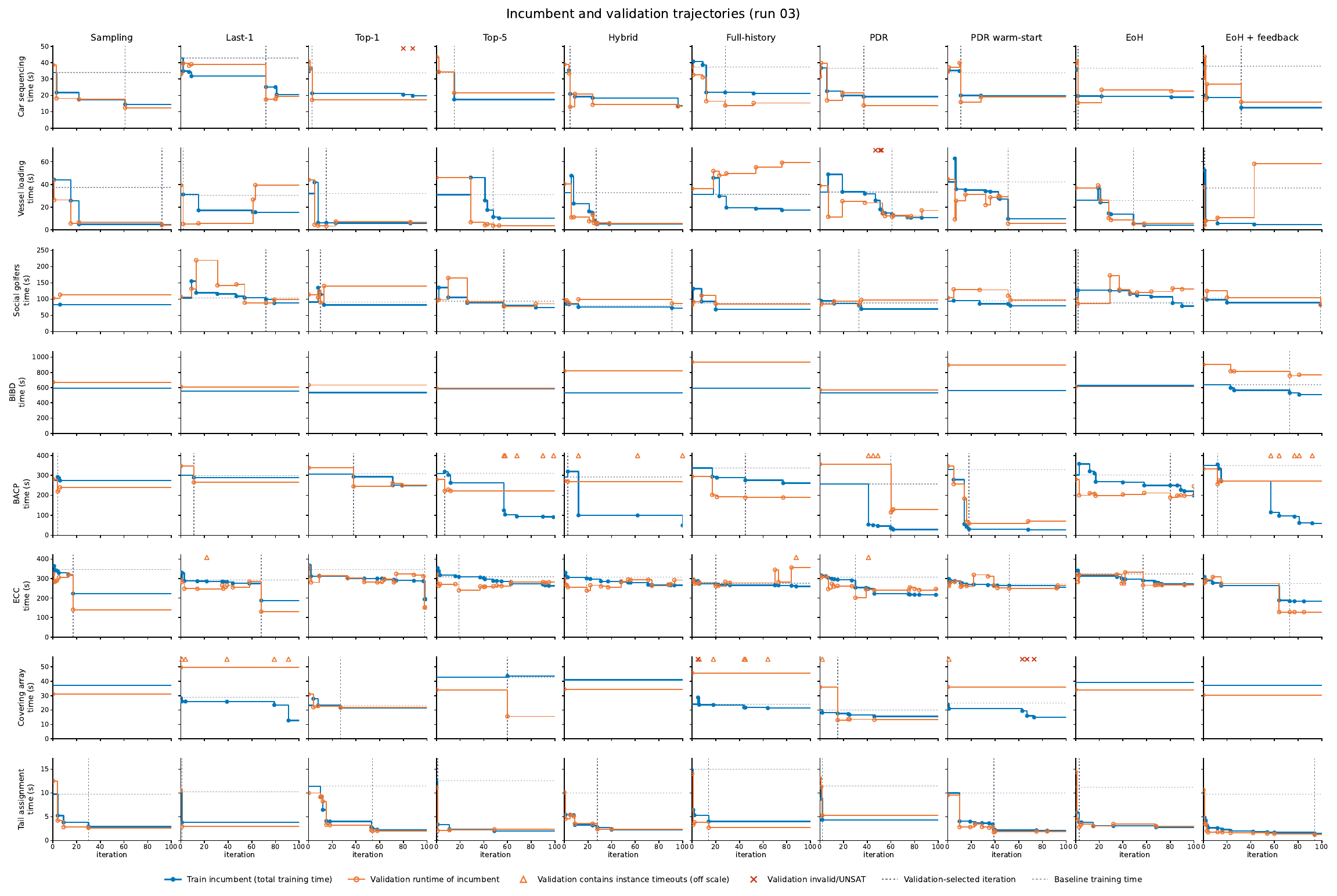}
\caption{Incumbent and validation trajectories, repetition 3.}
\label{fig:supp-grid-3}
\end{figure*}

Figures~\ref{fig:supp-grid-1}--\ref{fig:supp-grid-3} show incumbent and validation trajectories for every
problem--strategy cell. Validation totals containing per-instance PAR-2
timeout penalties are marked at the top.

\section{StreamLLM Adaptation Details}
\label{app:streamllm}

The StreamLLM adaptation compared in the main paper makes 20 calls, each
proposing five additive streamliners evaluated separately. A fresh call is
followed by two feedback calls carrying categorical outcomes; then the context
resets. The following initial instruction is exact and is followed by the same
CPMpy reference as the reformulation prompt.

\begin{lstlisting}[
  style=supplementlisting,
  caption={StreamLLM-single initial instruction.},
  label={lst:streamllm-initial-prompt}
]
Objective: Analyze the given CPMpy code and suggest five additional constraints to enhance the problem-solving process. These constraints can include streamlining, implied, symmetry-breaking, or dominance-breaking constraints.

Steps:

1. Analyze Content: Read the provided CPMpy code. Understand the problem being addressed, including its variables, constraints, and optimization goals.
2. Generate additional Constraints: Based on your analysis, create five unique constraints. These should offer targeted modifications or restrictions designed to reduce the search space effectively.
3. Always return your constraints as a JSON object, adhering to the structure: {"streamliner_1": "<CPMpy constraint expression>", ..., "streamliner_5": "<CPMpy constraint expression>"}. Your final output should exclusively be the JSON object containing the five constraints.
4. As a response, you will get feedback for each constraint. Some constraints might lead to errors, timeouts, or unsatisfiable instances.
5. Use the information provided in the previous step to generate five new, and hopefully better constraints.
6. Repeat steps 3 to 5 multiple times.

Compliance Rules:

1. Response Format: Your final output should exclusively be the JSON object containing the five constraints, adhering to the structure: {"streamliner_1": "<CPMpy constraint expression>", ..., "streamliner_5": "<CPMpy constraint expression>"}.
2. Code Quality: All CPMpy expressions must be syntactically correct and functional.
3. Creativity: You're encouraged to be innovative in proposing constraints, keeping in mind their purpose: to narrow down the search space efficiently without oversimplifying the problem.

CPMpy adaptation: Each value must be a single constraint expression that can be added to the existing model. For example, use an expression such as `cp.all([condition for i in range(n)])`, not Python statements such as `for i in range(n): model += ...`. Do not import modules, replace the solver, redefine variables, delete or mutate existing constraints, rewrite build_model, or return a full program.
\end{lstlisting}

Each feedback call uses this exact template:
\begin{lstlisting}[
  style=supplementlisting,
  caption={StreamLLM-single feedback instruction.},
  label={lst:streamllm-feedback-prompt}
]
Here are the results for the evaluation of the streamliners you provided:
- streamliner_1: <outcome>.
- streamliner_2: <outcome>.
- streamliner_3: <outcome>.
- streamliner_4: <outcome>.
- streamliner_5: <outcome>.
Based on this feedback, please give 5 new streamliners.
\end{lstlisting}

Proposals are AST-validated CPMpy expressions inserted into the unchanged
baseline. For StreamLLM-single, candidates are replayed in generation order;
a strict singleton incumbent is recorded whenever its summed solution-valid
training-time saving exceeds all earlier values. This yields 2--8 incumbents
per run (mean 4.25). Validation compares all incumbents with the baseline,
admits only streamliners solving every validation instance, and sends only the
winner to test without fallback. The baseline wins 3 of 24 runs. Four selected
models time out on one test instance each; none reports UNSAT, returns an
invalid answer, or errors on test. False-UNSAT outcomes occur before deployment
(4.6\% of training candidate--instance outcomes and 7 of 102 validation
incumbents) and are filtered by validation.

\paragraph{Portfolio diagnostic.}
The published protocol instead races three training-selected streamliners with
the baseline. Reconstructing completion times from the separately measured
per-instance runtimes, the ideal simultaneous runtime of that four-member
portfolio has an aggregate GM speedup of $1.87\times$. Granting PDR the same
baseline fallback, $T_{\mathrm{PDR+base}}(i)=\min\{T_B(i),T_{\mathrm{PDR}}(i)\}$
with $T_B(i)$ used whenever the PDR model returns no valid solution, gives
$2.41\times$, against $2.26\times$ for the strict single-model protocol
reported in the main paper. These reconstructions exclude process startup and
resource contention.

\FloatBarrier

\FloatBarrier

\end{document}